\documentclass[10pt,twocolumn,letterpaper]{article}

\usepackage[pagenumbers]{cvpr} 
\usepackage[accsupp]{axessibility}  

\definecolor{cvprblue}{rgb}{0.21,0.49,0.74}

\usepackage[pagebackref,breaklinks,colorlinks,allcolors=cvprblue]{hyperref}
\usepackage{multirow}
\usepackage{gensymb}
\newcommand{\prename}{\mbox{GeRaF}} 
\newcommand{\name}{\mbox{GeRaF 2.0}} 

\def\paperID{4045} 
\def\confName{CVPR}
\def\confYear{2026}

\title{Seeing through boxes: Non-Line-of-Sight 3D Reconstruction from Radar Signals}

\author{%
  Jiachen Lu\textsuperscript{*}, \quad  Hailan Shanbhag\textsuperscript{*}, \quad Haitham Al Hassanieh\\
École Polytechnique Fédérale de Lausanne  (EPFL)
}

\begin{document}
\maketitle
\def\thefootnote{*}\footnotetext{Co-primary first authors, indicates equal contribution.}
\def\thefootnote{\dag}\footnotetext{Project page: \url{https://sens.epfl.ch/research/geraf/}}
\begin{abstract}
Reconstructing object geometry from radio frequency (RF) signals is fundamentally challenging due to the lensless imaging nature of RF sensing, which leads to low spatial resolution and high noise. 
Unlike light signals, RF signals can penetrate occlusions and thus capture information about hidden scenes. 
Existing Non-Line-of-Sight (NLoS) 3D neural reconstruction methods can recover coarse surfaces inside enclosed environments but often suffer from unstable optimization, noisy surface geometry, and surface ambiguity, failing to produce accurate zero-level sets from the signed distance field (SDF).
These limitations largely stem from neglecting the role of Line-of-Sight (LoS) geometry outside the enclosed region, which provides valuable physical constraints for modeling signal propagation.
In this paper, we introduce a Unified LoS and NLoS neural geometry reconstruction framework \name{} that leverages the outside LoS geometry to model and guide RF propagation from the LoS region into the NLoS region. 
By integrating visual LoS priors into the neural field formulation, \name{} achieves stable training and physically consistent reconstruction of both visible and hidden geometry, setting a new state-of-the-art in RF-based geometry reconstruction.
\end{abstract}
    
\section{Introduction}
\label{sec:intro}
 
Radio frequency (RF) reconstruction has exploded in recent years as a robust and versatile sensing modality due to its unique ability to see \textit{through} occlusions and remain reliable under challenging visibility conditions. 
This unique ability to see through occlusions and operate in non-line-of-sight scenarios while being safe for humans~\cite{wu2015safe}, unlocks a large range of applications, such as allowing robots to see hidden objects inside boxes or behind clutter or allowing smart home devices to interact with occluded regions~\cite{adib2013see,yue2020bodycompass}.

However, directly reconstructing 3D objects from RF signals is challenging.
Due to their \textit{lensless} nature, the pinhole camera model commonly used in vision does not apply.
Each antenna can receive signals from the whole scene, bringing challenges in high-cost sampling, low resolution, and high noise levels. 
Traditionally, it is common to use an array of antennas and combine the signals across antennas~\cite{yanik2019near}, producing more interpretable reconstructions. 
However, compared to vision reconstruction, the reconstruction from RF is corrupted by noise artifacts, missing surface patches due to specular RF reflections, and very low spatial resolution due to physical limitations of the antenna array apertures.

\begin{figure}[t]
    \centering
    \includegraphics[width=\linewidth]{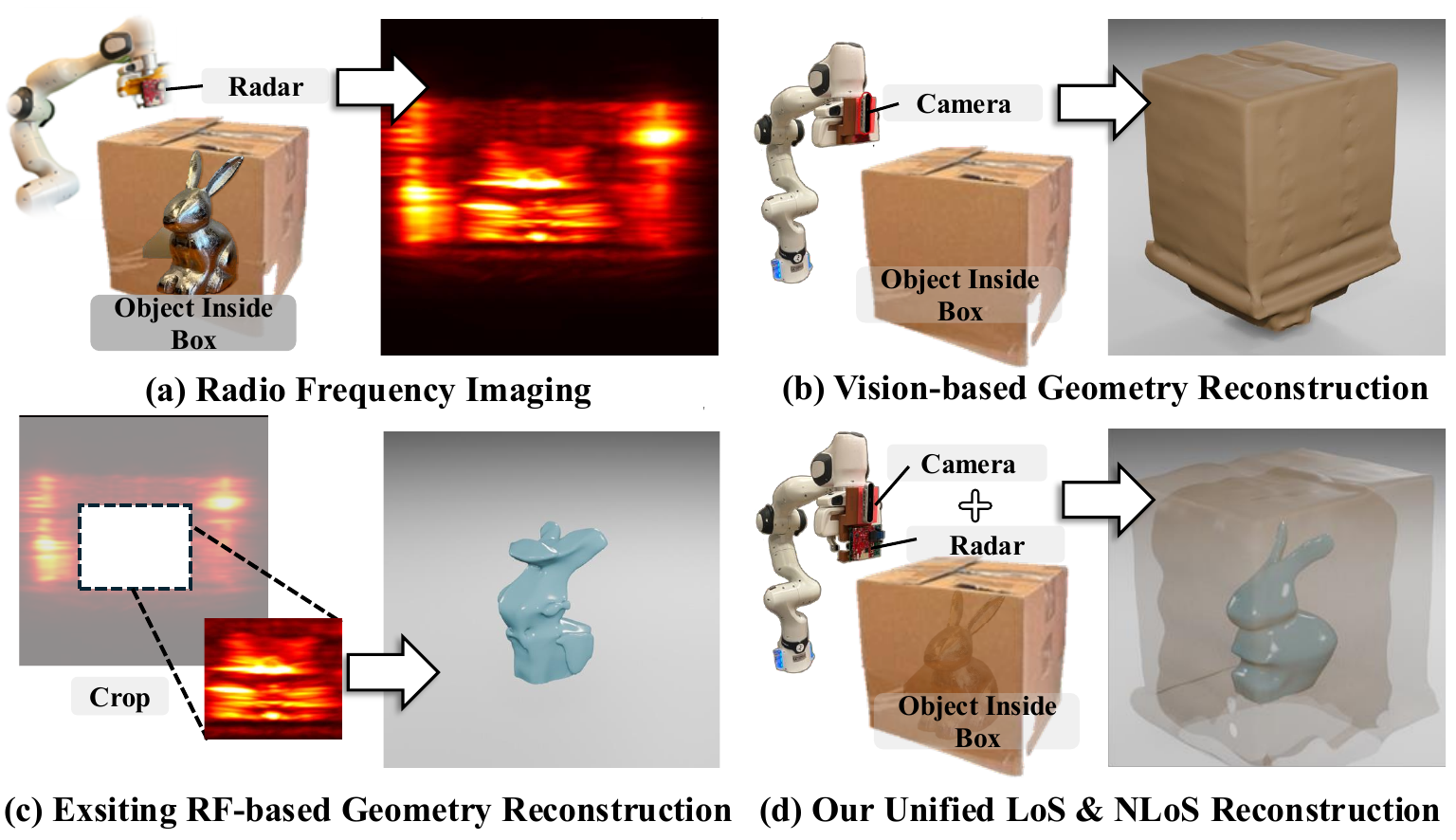}
    \caption{This is the first work which uses line-of-sight modalities to \textit{boost} non-line-of-sight 3D reconstruction for high resolution 3D reconstruction. \textbf{a)} Radar heatmap from combining multi-view images, \textbf{b)} vision based reconstruction of the line-of-sight surface, \textbf{c)} GeRaF~\cite{lu2025geraf} which crops the box out of reconstruction, and \textbf{d)} \name{} which combines line-of-sight and non-line-of-sight reconstruction.}
    \label{fig:teasor}
\vspace{-1em}
\end{figure}

Recently, there has been a growing interest in neural reconstruction methods for radio frequency~\cite{borts2024radar, huang2024dart, zhang2025rf4d, kung2025radarsplat, lu2025geraf},
which try to adapt and apply these methods to RF sensing to represent the geometry of a scene continuously, enabling smoother and more accurate 3D representations.
Many of these works~\cite{borts2024radar, huang2024dart, zhang2025rf4d, kung2025radarsplat} only reconstruct scenes in {\bf Line-of-Sight (LoS)} and are geared towards mapping the environment for autonomous navigation rather than reconstructing detailed 3D models of complex objects.
Our recent work \prename{}~\cite{lu2025geraf}, reconstructs surfaces in \textbf{Non-Line-of-Sight (NLoS)} (e.g. behind paper or in a box)
by simply cropping out the occlusion and treating as if it was not there (i.e. treating LoS and NLoS in the same way).
However, the LoS model incorrectly assumes that the wireless signal that reaches the NLoS surface after passing through the occluding surface, is completely unaffected by the LoS surface.  
In reality, interactions with the LoS occlusion is unavoidable, because the wireless signal that passes through the LoS occlusion is partially \textit{reflected} and \textit{attenuated} by the occluding surface~\cite{stone1997electromagnetic, dhekne2018liquid}. 

As a result, \prename{}~\cite{lu2025geraf} suffers from:  
(1) inaccurate surface reconstruction, since wireless reflections from the visible surfaces can ``leak" into the hidden region, appearing as noise that alters the reconstruction, as shown in Fig.~\ref{fig:teasor}(c), where the bunny has a strange ``hat" shape influenced by the box;  
(2) unstable training, because different LoS geometries (e.g., boxes of different shapes and sizes) can alter the optimization landscape, leading to failure to converge to a stable surface in some cases;  
(3) surface ambiguity, since the LoS geometry affects the power of the wireless signal reaching the NLoS regions, it becomes difficult to normalize the signal strength and, as a result, to determine the true surface (i.e., the zero-level set of the SDF). For example, the surface in Fig.~\ref{fig:teasor}(c) is not selected based on the SDF being equal to zero, but rather offset by a few centimeters.
On the other hand, the vision-based reconstruction methods well known from~\cite{mildenhall2021nerf, wang2021neus} are advantageous for stable training and accurate surface recovery. 
As shown in Fig.~\ref{fig:teasor}(b), vision successfully reconstructs a highly accurate LoS box, but it cannot see inside the box.
This prompts us to ask the question: \textit{can we leverage \textbf{visible} information outside the boxes to help see through them?}

To address this, we propose a \textbf{Unified Line-of-Sight and Non-Line-of-Sight (ULoS)} neural geometry reconstruction framework \textbf{\name{}} that exploits stable and accurate information from line-of-sight modalities outside the box to guide low-resolution and noisy non-line-of-sight modalities inside the box.
(1) We represent the combined LoS and NLoS regions as nested, closed, and compact sets, enabling a unified representation within a single field named the ULoS Signed Distance Field (ULoS SDF).
This formulation supports consistent optimization across both regions and helps mitigate LoS-induced artifacts within the NLoS area.
(2) We then propose a ULoS Rendering technique, which incorporates the vision-pretrained SDF to provide a stable initialization for training the ULoS SDF with RF signals.
(3) To address the surface ambiguity problem and determine the correct zero-level set of the ULoS SDF, we introduce a second stage of training which aligns the vision-pretrained SDF and the RF-trained ULoS SDF {\bf\em outside} the box, and prove that this alignment transfers to alignment {\bf\em inside} the box.
With these three components, Fig.~\ref{fig:teasor}(d) shows that \name{} achieves global surface reconstruction across both LoS and NLoS regions, producing a clean and accurate surface with the correct zero level of the SDF.

We evaluate \name{} using a 77\,GHz mmWave radar mounted on a Franka Research 3 robotic arm, imaging a variety of real-world objects inside different boxes.
Our results demonstrate that \name{} outperforms previous works and takes a significant step towards more robust and accurate 3D reconstruction from RF signals behind occlusions.

\section{Related Work}
\label{sec:related}
\noindent\textbf{Vision-Based Neural 3D Reconstruction:} 
Neural Radiance Fields (NeRF)~\cite{mildenhall2021nerf} and Gaussian Splatting~\cite{kerbl20233d} introduced using learnable parameters to reconstruct 3D scenes from multi-view images, which creates a neural implicit representation of the 3D scene. 
Motivated by this~\cite{niemeyer2020differentiable, yariv2020multiview, yariv2021volume, wang2021neus,huang20242d, jiang2024gaussianshader, liang2024gs, gao2024relightable, chen2023neusg}, more works further separates scene components into explicit geometric representations and reconstructs detailed surface meshes.
\cite{srinivasan2021nerv, boss2021nerd, munkberg2022extracting, yao2022neilf, jin2023tensoir, liu2023nero, gu2025irgs, yao2025reflective} take it a step further to predict lighting and material properties by learning complicated light interactions. They base the reconstruction on the explicit forward rendering equation adding in the reflectance distribution function.
However, all of these 3D representations are based on optical signals, which are not easily translatable to radio frequency.

Further more there are works that do NLOS using optical sensors~\cite{o2018confocal,liu2019non,lindell2019wave,velten2012recovering}, however, they tackle around-the corner imaging, requiring the signal to reflect off of surfaces, whereas our problem looks at wireless signals penetrating through surfaces. 

\noindent\textbf{3D Radar Reconstruction:}
Deep learning techniques have been used for imaging or point cloud completion in the context of self-driving cars. However, these works are geared towards street-level scenes and LoS (unoccluded) large objects like cars and pedestrians~\cite{hussein20253d, guan2020through, sun20213drimr, lai2024enabling, sun20223d, sun2022r2p}.

mmNorm~\cite{dodds2025non} performs non-line-of-sight surface reconstruction by estimating a normal field and optimizing over isosurfaces obtained by inverting the normal field. However, unlike our system, mmNorm focuses on one-sided 3D reconstruction (front view instead of $360^{\circ}$) and, similar to other works, it simply crops the occlusion out.

\noindent\textbf{RF Neural Implicit Reconstruction:} ~\cite{borts2024radar,huang2024dart, zhang2025rf4d, kung2025radarsplat} try to reconstruct self-driving car scenes by rendering a power distribution of the wireless signal and learning the occupancy of different locations in the scene. Another set of work~\cite{barbierrenard2025multiview3dsurfacereconstruction, tan2024fast} apply neural implicit reconstruction to satellite images. However, all of these works perform reconstruction specifically tailored for reconstructing \textit{large} scale scenes such as streets or satellite images and don't address close-range high-resolution object reconstruction, which requires different wireless propagation modeling as explained in the appendix. 

For near range reconstruction, authors of~\cite{takawale2025spinr}, propose a method for 3D neural reconstruction of objects. However, their evaluation is limited to simulated data, which cannot represent the complexity of real world experiments with wireless signals. 
Most recently, GeRaF\cite{lu2025geraf} propose a neural reconstruction method, tested on real world experiments, specifically for near-field objects, by using a physical rendering model of radio signals to learn a surface model. However, GeRaF avoids the need to model occlusions (eg. boxes, paper) by simply cropping the reflections from the occlusion out of the radar image, which introduces additional noise and degrades surface reconstruction.

\noindent\textbf{Radar-Vision Joint Perception:} A plethora of works have explored the benefits of combining radar and camera perception~\cite{wu2023mvfusion,el2015toward,lin2024rcbevdet,chen2024towards,yang2025zfusion,ali2019multi,yu2023sparsefusion3d}. However, these papers are geared for self-driving car scenarios and bounding box detection, not reconstruction. 
Moreover, none of these works have used radar-camera fusion for neural implicit reconstruction for high-resolution 3D reconstruction.
RadarSim~\cite{chenradarsim} proposes a camera-radar joint reconstruction framework to create novel Doppler-range images from camera initialized 3D geometry for improvedd radar simulation. However, their goal is accurate radar synthesis from line-of-sight conditions, while our works tackles a different challenge: non-line-of-sight reconstruction from camera-radar joint observations.

\section{Wireless Technical Background}
\label{sec:background}
\subsection{Radio Frequency Background}
\noindent \textbf{Waveform}
\label{sec:background-mm}
A radar transmits a wireless
waveform and receives reflections that come from the signal bouncing off of various objects in the environment.
Our system uses Frequency Modulated Continuous Wave (FMCW) and antenna arrays to resolve range, azimuth and elevation ambiguity as a result of the lensless nature of wireless signals.
The received signal is multiplied with the conjugate of the transmitted signal and is expressed as:
\begin{equation}
\label{eq:fmcw}
s(t) =  A \cdot e^{-j 2 \pi(\nu+kt)   d /c}  = A \cdot e^{-(j 2\pi k \tau )t} \cdot e^{-j 2\pi \nu \tau} 
\end{equation}
where $A$ is the signal amplitude, $d$ is the round-trip propagation distance, $c$ is the speed of light, $\tau=d/c$ is the round-trip delay, $\nu$ is the starting frequency, and $k$ is the slope of the frequency change. For multiple reflectors in the scene, we receive the linear combination of Eq.~\ref{eq:fmcw}.

\noindent \textbf{Reflector Interaction}
Unlike light, whose short wavelength causes diffused reflections, RF signals have much longer wavelengths, making most surfaces appear smooth and produce primarily specular reflections~\cite{richards2010principles}.
In this paper we follow the reflection model as used in ~\cite{lu2025geraf}.
Given an input signal with amplitude $ A_{\text{TX}} $, the received amplitude $ A_{\text{RX}} $ is expressed as:
{\setlength{\abovedisplayskip}{6pt}%
 \setlength{\belowdisplayskip}{6pt}%
\begin{equation}
\label{eq:reflection_angle}
A_{\text{RX}} \propto \frac{a}{(4\pi u)^2} A_{\text{TX}}\, (\omega_o \cdot \omega_r)
\end{equation}}
where \( a \) is the reflectivity, \( u \) is the propagation distance from the reflection point to the receiver,
\( \omega_r \) is the incoming vector of the RF signal, and \( \omega_o \) is the outgoing vector.

\subsection{Lensless Volumetric Rendering}
\noindent \textbf{Signal Tracing}
While vision-based rendering relies on optical lenses to filter and focus relevant light rays, wireless sensing operates through \textit{lensless imaging}, capturing all incoming RF signals without directional filtering. 
The RF signal received at time $t$ by an antenna located at $\mathbf{x}_{\text{ant}}$ can be expressed as:
\begin{equation}
\label{eq:ref_sim}
s(\mathbf{x}_{\text{ant}}, t, u) = \sum_{\mathbf{x} \in \Omega_\text{ULoS}} A_{\text{rx}}(\mathbf{x}) \, e^{-j 2\pi k \tau_{\mathbf{x}} t} \, e^{-j 2\pi \nu \tau_{\mathbf{x}}},
\end{equation}
where $\tau_{\mathbf{x}}$ denotes the propagation delay from point $\mathbf{x}$ to the antenna, and $A_{\text{rx}}(\mathbf{x})$ represents the received amplitude at $\mathbf{x} = \mathbf{x}_\text{ant} + \omega_r \cdot u$, which can be modeled using Eq.~\ref{eq:reflection_angle} as:
\begin{equation}
\label{eq:signal_amp_tracing}
A_{\text{rx}}(\mathbf{x}_{\text{ant}}, \omega_r, u) 
= \frac{\mathbf{a}(u)}{(4\pi)^2} \, (\omega_o \cdot \omega_r) \, T(u)^2 \, \rho(u) \, \mathbf{A}_{\text{tx}} \, \mathrm{d}t,
\end{equation}
where \( \mathbf{a}(u) \) is the reflectivity,  
\( \omega_o \) is the outgoing vector computed as  
\(
\omega_o = \omega_i - 2(\mathbf{n} \cdot \omega_i)\mathbf{n},
\) 
with \( \omega_i \) as the incoming signal vector and \( \mathbf{n} \) the surface normal.  
Here, \( \mathbf{A}_{\text{tx}} \) denotes the equivalent transmitted power.  
The terms \( T(u) \) and \( \rho(u) \) are adopted from volumetric rendering in vision~\cite{wang2021neus}, where \( T(u) \) represents accumulated transmittance and \( \rho(u) \) denotes opacity.  
Since radar sensing involves two-way propagation, the transmittance term \( T(u) \) is squared.

\noindent \textbf{Lensless Sampling and Lensless Rendering}
Each antenna receives signals from all incoming directions, thus the lensless sampling strategy introduced in
\cite{lu2025geraf} is used to avoid exhaustive grid-based sampling for every antenna; the computation of opacity and accumulated transmittance is shared across antennas.
Specifically, the quantities $\rho(u)$ and $T(u)$ only need to be computed once for each spatial position, as they are shared by all antenna rays that intersect the same voxel.

As shown in Fig.~\ref{fig:alpha-blending}, lensless sampling begins by casting parallel rays aligned with the radar’s primary direction $\omega_p$ from the antenna aperture.  
The opacity $\rho(u)$ and transmittance $T(u)=\exp{\left(-\int_0^u\rho(v)dv\right)}$ is computed along the primary ray using traditional rendering~\cite{wang2021neus}.
The true transmittance \( T(u') \) along a real ray (orange ray in Fig.~\ref{fig:alpha-blending}) pointing toward the antenna is avoided from being recomputed.
Instead, the sigmoid-SDF values are adjusted at the scene boundary \( \partial \Omega_{\mathrm{ULoS}} \) (green and orange points in Fig.~\ref{fig:alpha-blending}):
\begin{equation}
\label{eq:lens-less-transmittance}
T(u') = T(u) - \Phi_s(f(\mathbf{x}(u_s))) + \Phi_s(f(\mathbf{x}(u'_s))),
\end{equation}
where $\Phi_s(\cdot)$ denotes the sigmoid function, and $\mathbf{x}(u_s)$ and $\mathbf{x}(u'_s)$ represent the starting points of the primary and secondary rays.

\begin{figure}[tb]
    \centering
    \includegraphics[width=\linewidth]{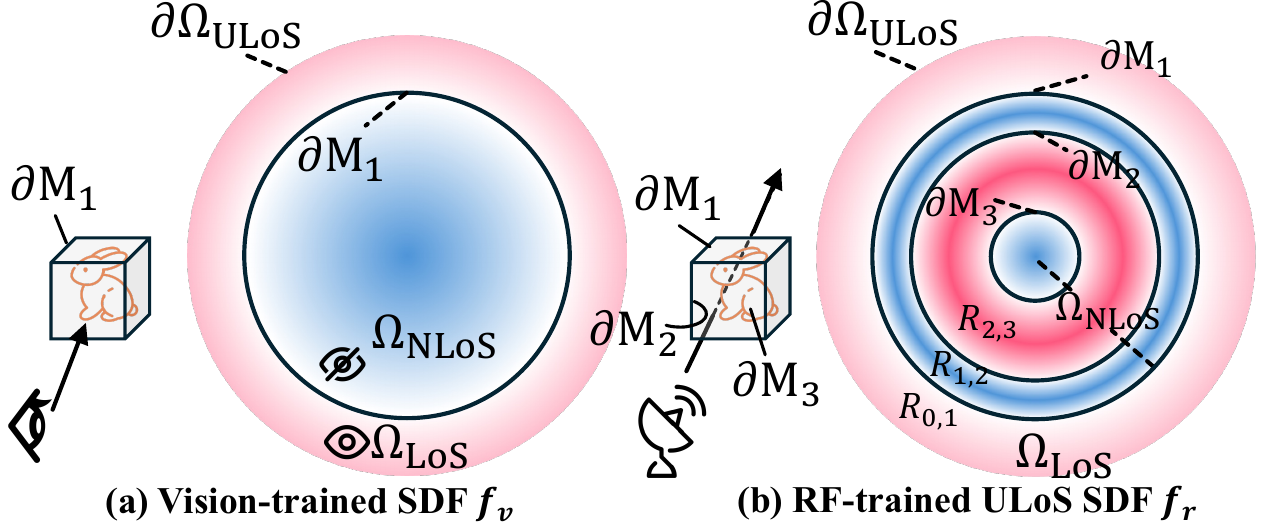}
    \caption{
\ \textbf{a)} Vision-trained SDF uses negative values (blue) and positive values (red) to separate the inside and outside.
\ \textbf{b) }RF-trained ULoS SDF models the scene as a series of nested closed and compact sets. Areas with a strong radio frequency interaction (e.g., carton, metal) (blue) are assigned negative values, weak interactions (e.g., air) regions (red) are assigned positive values.
}
    \label{fig:ulos_sdf}
\end{figure}
\section{Rendering from Radio Frequency Signals}
\subsection{ULoS Scene Representation}
\label{sec:scene-rep}
As shown in Fig.~\ref{fig:ulos_sdf}(a),  
according to visibility conditions, the entire scene can be split into a 
line-of-sight (LoS) region and a non-line-of-sight (NLoS) region, denoted by two closed and compact sets:  
$\Omega_{\mathrm{LoS}}\subset \mathbb{R}^3$ and 
$\Omega_{\mathrm{NLoS}}\subset \mathbb{R}^3$. 
The union of these two regions defines the Unified Line-of-Sight and Non-Line-of-Sight (ULoS) domain:
\begin{equation}
    \Omega_{\mathrm{ULoS}}
    = \Omega_{\mathrm{LoS}} \cup \Omega_{\mathrm{NLoS}}.
\end{equation}

Fig.~\ref{fig:ulos_sdf}(b) illustrates the ULoS scene.   
We represent the scene as a series of nested, closed, and compact sets in three-dimensional Euclidean space. 
Let $M_1, \dots, M_n \subset \mathbb{R}^3$ be such that
\begin{equation}
\label{eq:subsets}
    M_n \subset \operatorname{int}(M_{n-1})
    \subset \cdots 
    \subset \operatorname{int}(M_1)
    \subset \operatorname{int}(\Omega_{\mathrm{ULoS}}),
\end{equation}
where $\operatorname{int}(\cdot)$ denotes the interior of a set.

The spatial region between two consecutive layers $M_i$ and $M_{i+1}$ is defined as
\begin{equation}
    R_{i,i+1} 
    = \left\{ \mathbf{x} \in \mathbb{R}^3 \,\middle|\, \mathbf{x} \in \operatorname{int}(M_i),\,
        \mathbf{x} \notin \operatorname{int}(M_{i+1}) \right\},
\end{equation}
The outermost region is defined by
\[
    R_{0,1} = 
    \{\mathbf{x} \in \mathbb{R}^3 
      \mid 
      \mathbf{x} \in \Omega_{\mathrm{ULoS}},\,
      \mathbf{x} \notin \operatorname{int}(M_1)\}.
\]
The boundary of each intermediate region is the union of its two layer surfaces:
\begin{equation}
    \partial R_{i,i+1} = \partial M_i \cup \partial M_{i+1}.
\end{equation}

The observer is set outside the outermost domain $\Omega_\mathrm{ULoS}$. From Eq.~\eqref{eq:subsets}, the outermost set $M_1$ fully contains all inner subsets.
Therefore, its boundary $\partial M_1$ blocks visible light and defines the limit of optical visibility.  
Consequently, the line-of-sight region is given by
\(
    \Omega_{\mathrm{LoS}} = R_{0,1},
\)
while the non-line-of-sight region corresponds to all nested layers within the box:
\(
    \Omega_{\mathrm{NLoS}} = \bigcup_{i=1}^{n} M_i = M_1.
\)

\paragraph{ULoS SDF}
Vision-based neural geometry reconstruction methods~\cite{wang2021neus, muller2022instant} represent scenes using a Signed Distance Function (SDF) $f_v$, which clearly distinguishes between the inside, outside, and surface of objects.  
However, RF signals interact with the environment in fundamentally different ways.  
The traditional binary notion of ``inside'' and ``outside'' becomes ambiguous, especially across multiple layers, as NLoS regions can still be partially transparent to RF signals.  
To address this, we introduce a unified Signed Distance Function tailored for both LoS and NLoS RF propagation, denoted as the \emph{ULoS SDF}.  
We denote this RF-specific distance field as $f_r$ throughout our formulation.

We define the ``negative" (or interior) region for a given medium as the set of points that impose 
a strong radio frequency interaction against the propagating field, and the ``positive" (or exterior) region 
as the set of points with a weak radio frequency interaction.
For example, Fig.~\ref{fig:ulos_sdf}(b) shows a simple case of an object inside a box.  
The scene contains three surfaces: the outer surface of the box, the inner surface of the box, and the surface of the object.  
The region between the box surfaces and the interior of the object is assigned negative values, while all other regions are positive.

\begin{figure}[tb]
    \centering
    \includegraphics[width=\linewidth]{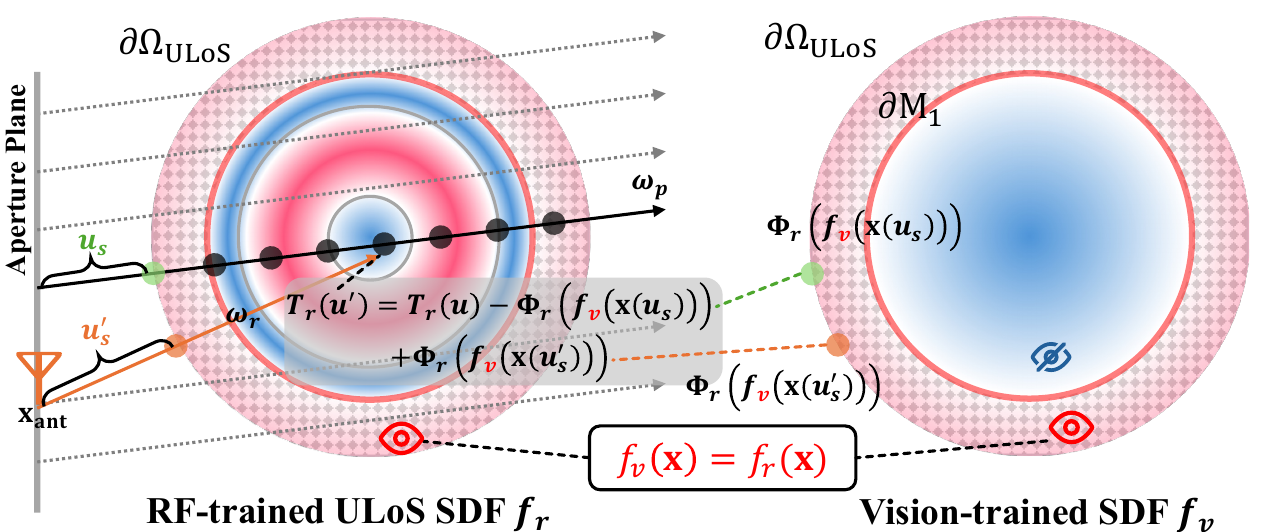}
    \caption{
Lensless sampling and ULoS lensless rendering. 
Gray rays represent the primary sampling direction and orange ray indicates the actual ray pointing from the antenna.  
ULoS lensless rendering: we exploit the fact that vision-trained SDF and RF-trained ULoS SDF share identical values in the LoS region (shaded), allowing us to to adjust the accumulated transmittance.
}
\vspace{-1em}
    \label{fig:alpha-blending}
\end{figure}
Formally, the sign of the ULoS SDF is determined by the relative interference coefficient:
\[
    \operatorname{sign}(f(\mathbf{x})) =
    \begin{cases}
        -1, & \text{if } R_{i,i+1} \text{ is strong interaction}, \\[3pt]
        +1, & \text{if } R_{i,i+1} \text{ is weak interaction}.
    \end{cases}
\]

To maintain continuity and geometric consistency across multi-layered boundaries,
the signed distance for each region $R_{i,i+1}$ is defined as the \textit{minimum Euclidean distance to its nearest bounding surfaces}:
\begin{equation}
    f(\mathbf{x}) =
    \begin{cases}
        \operatorname{sign}(f(\mathbf{x})) 
        \min\!\big(
            d(\mathbf{x}, \partial M_i), \\[2pt]
            \qquad d(\mathbf{x}, \partial M_{i+1})
        \big), 
        & \mathbf{x} \in \operatorname{int}(R_{i,i+1}), \\[5pt]
        0, & \mathbf{x} \in \partial R_{i,i+1},
    \end{cases}
\label{eq:medium_sdf_compact}
\end{equation}
where $d(\mathbf{x}, \partial M_i)$ is the Euclidean distance from $\mathbf{x}$ to the surface $\partial M_i$.
This unified formulation preserves geometric continuity while maintaining physically meaningful sign semantics for each modality.

\begin{figure*}[htb]
    \centering
    \includegraphics[width=\linewidth]{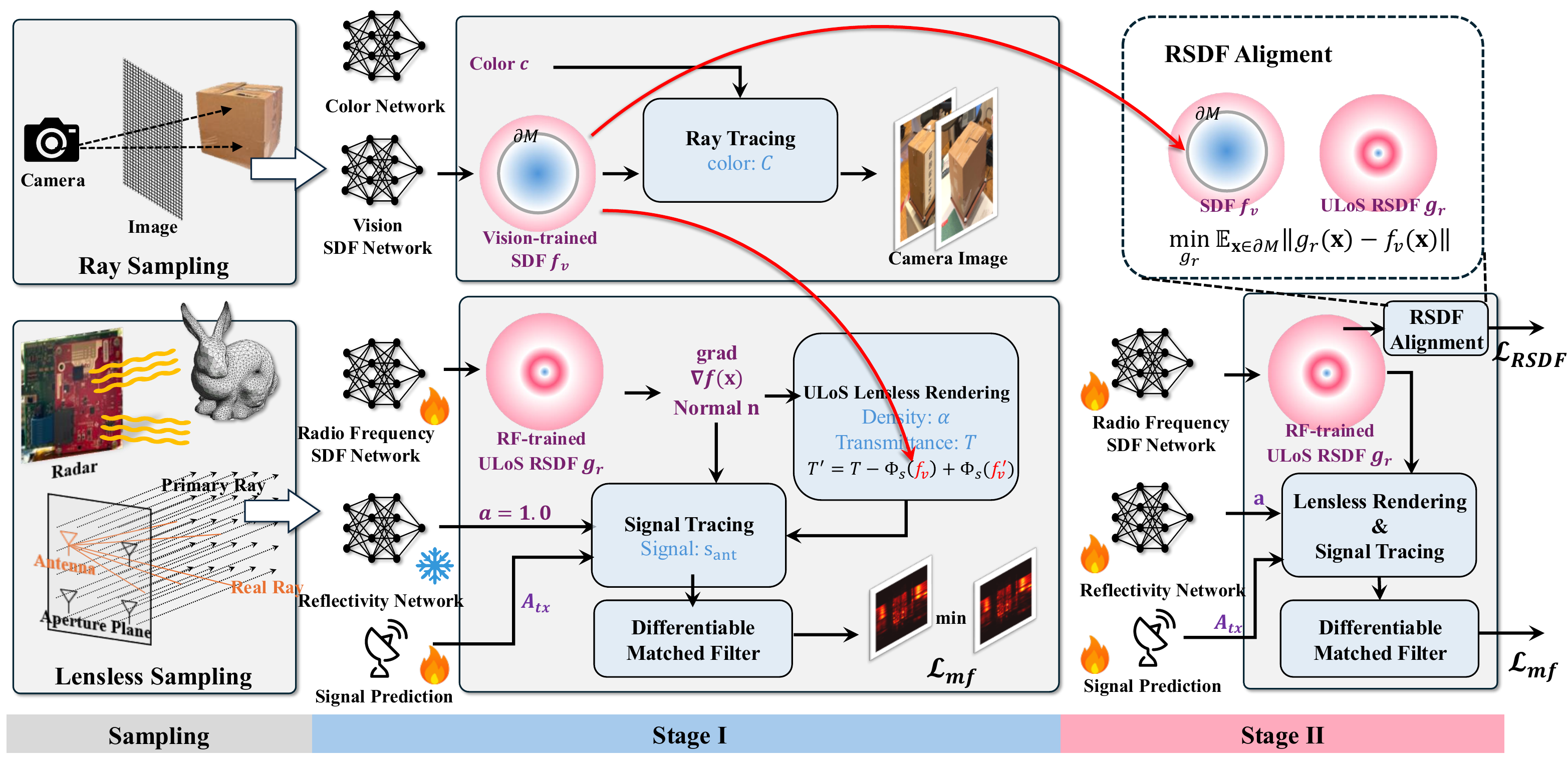}
    \caption{
Overall pipeline of \name{}.  
{\bf\em Top:} The vision-pretrained SDF on the outside of the box.  
{\bf\em Bottom:} The training pipeline for RF signals. The pipeline begins with lensless sampling.  
In the first stage of training, we freeze the Reflectivity Network and use the vision-pretrained SDF to adjust transmittance in the ULoS Lensless Rendering module.  
In the second stage, we use the vision-pretrained SDF to align the relative SDF, thereby addressing the surface ambiguity problem.
}
    \vspace{-1.25em}
    \label{fig:pipeline}
\end{figure*}
\subsection{ULoS Lensless Rendering}
\label{sec:ulos-rendering} 
It is important to note that the terms $\Phi_s(f_{\text{NLoS}}(\mathbf{x}(u_s)))$ and $\Phi_s(f_{\text{NLoS}}(\mathbf{x}(u'_s)))$ in Eq.~\ref{eq:lens-less-transmittance} are excluded from the backpropagation process due to their high computational cost.  
This omission, however, introduces errors in gradient propagation, leading to suboptimal optimization.  
Moreover, these terms exhibit a significant bias during network initialization, resulting in skewed transmittance adjustment in the early stages of training.

This problem is particularly challenging in NLoS scenes, where limited signal visibility makes stable optimization difficult.  
However, in the ULoS setting, additional geometric information from the LoS domain can play a crucial role.  
As illustrated in the shaded region of Fig.~\ref{fig:alpha-blending}, the \textit{vision-trained SDF and the RF-trained ULoS SDF share identical values outside the box region \( R_{0,1} \)}:
{\setlength{\abovedisplayskip}{6pt}%
 \setlength{\belowdisplayskip}{6pt}%
\begin{equation}
\label{equ: obs3}
    f_v(\mathbf{x}) = f_r(\mathbf{x}), 
    \qquad \mathbf{x} \in R_{0,1}.
\end{equation}}
Since the starting points of the primary rays lie in free space outside the enclosing box, the SDF and the ULoS SDF are equivalent at these locations.  
By initializing the ULoS SDF at these starting points using a well-converged, vision-pretrained neural reconstruction, we provide a stable prior for training with RF signals, leading to faster convergence and improved consistency.

\subsection{Overall Pipeline}
The overall pipeline is illustrated in Fig.~\ref{fig:pipeline}. We perform reconstruction in the following way:
\textbf{(1)} We train the SDF of the exterior of LoS surface using NeuS~\cite{wang2021neus}. After training, the vision model is fixed for the remainder of the pipeline.
\textbf{(2)} Then we move onto RF reconstruction, and begin with lensless sampling strategy to generate point samples in space.
\textbf{(3)} Each sampled point is processed by three sub-networks: the SDF Network, the Reflectivity Network, and the Signal Power Prediction Network. The SDF Network is initialized with the vision-pretrained SDF and is used to predict the ULoS SDF, which is used to compute opacity and transmittance.  
The Reflectivity and Signal Power Networks estimate surface reflectance and received signal power, respectively.
\textbf{(4)} From the outputs of the sub-networks, the ULoS Lenless Rendering module (Sec.~\ref{sec:ulos-rendering}) simulates the received antenna signal. Within this module, the vision-pretrained SDF is used to adjust the transmittance rather than relying on the model under training.
\textbf{(5)} Finally, a differentiable matched filter is applied to render the matched filter heatmap, and the loss is computed between the rendered and ground-truth heatmaps.

However, after this pipeline, the reconstructed surface still suffers from inaccuracies and an incorrect zero-level set of the SDF.  
This issue, known as the surface ambiguity problem~\cite{dodds2025non}, arises from the inherent difficulty of normalizing radar signal strength.  
To address this, we introduce a second-stage training process, described in detail in Sec.~\ref{sec:surface_ambiguity}.

\section{The Surface Ambiguity Problem}
\label{sec:surface_ambiguity}
The surface ambiguity problem~\cite{dodds2025non} arises from the inherent difficulty of normalizing the radar signal strength.
Unlike RGB images in LoS scenes, where light intensity is pre-normalized to the range $[0,1]$, radar signals cannot be normalized in the same way due to their strong dependence on scene geometry, material properties, and signal attenuation. 
As a result, the reflectivity, predicted signal power, and the zero-level surface of the SDF become mutually entangled variables. 
Without additional constraints or external information, the network cannot uniquely determine the correct surface configuration, leading to an incorrect zero-level SDF definition and inaccurate surface geometry.

\subsection{Relative Signed Distance Function}

To facilitate analysis, we adopt the \textit{relative signed distance function} (RSDF)~\cite{dodds2025non}, 
denoted by $g(\mathbf{x})$, which is defined as the SDF offset by some unknown constant.  
The gradient of the RSDF is identical to that of the conventional signed distance function (SDF) $f(\mathbf{x})$:{\setlength{\abovedisplayskip}{6pt}%
 \setlength{\belowdisplayskip}{6pt}%
\begin{equation}
    \nabla f(\mathbf{x}) = \nabla g(\mathbf{x}),
    \qquad \forall\, \mathbf{x} \in \Omega.
    \label{eq:rsdf_grad}
\end{equation}}
Our objective is for the RSDF to converge to the true SDF, i.e.,
\[
    f(\mathbf{x}) = g(\mathbf{x}), 
    \qquad \forall\, \mathbf{x} \in \Omega.
\]
To formalize this equivalence, we present the following proposition.

\noindent\textbf{Proposition.}
Let $f, g : \Omega \to \mathbb{R}$ be two continuously differentiable scalar fields defined on a connected region 
$\Omega \subset \mathbb{R}^3$.  
If the following two conditions hold:
\begin{enumerate}
    \item $\nabla f(\mathbf{x}) = \nabla g(\mathbf{x})$ for all $\mathbf{x} \in \Omega$, and
    \item $f(\mathbf{x}) = g(\mathbf{x})$ for all $\mathbf{x}$ on a closed surface $S \subset \Omega$,
\end{enumerate}
then
\[
    f(\mathbf{x}) \equiv g(\mathbf{x}), 
    \qquad \forall\, \mathbf{x} \in \Omega.
\]

\noindent\textbf{Proof.}
Define $h = f - g$.  
Since $\nabla f = \nabla g$, it follows that $\nabla h = \mathbf{0}$ for all $\mathbf{x} \in \Omega$.  
A differentiable scalar field with zero gradient on a connected domain must be constant; 
hence $h = C$ for some $C \in \mathbb{R}$.  
From condition~(2), $f(\mathbf{x}) = g(\mathbf{x})$ for all $\mathbf{x} \in S$, 
implying $C = 0$.  
Therefore, $h(\mathbf{x}) = 0$ for all $\mathbf{x} \in \Omega$, 
and thus $f \equiv g$ throughout $\Omega$.

Condition (1) is directly satisfied by the definition of the RSDF. 
Therefore, it remains to ensure that the RSDF derived from radar signals, denoted by $g_r(\mathbf{x})$, 
matches the corresponding SDF, $f_r(\mathbf{x})$, on a closed reference surface 
$S \subset \Omega_{\mathrm{ULoS}}$.

\noindent\textbf{Observation.}
Within the LoS region $R_{0,1}$, the RF-trained SDF and vision-trained SDF coincide, 
i.e., $f_r(\mathbf{x}) = f_v(\mathbf{x})$.  

\subsection{Relative Signed Distance Function Alignment}

\begin{figure}[t!]
    \centering
    \includegraphics[width=\linewidth]{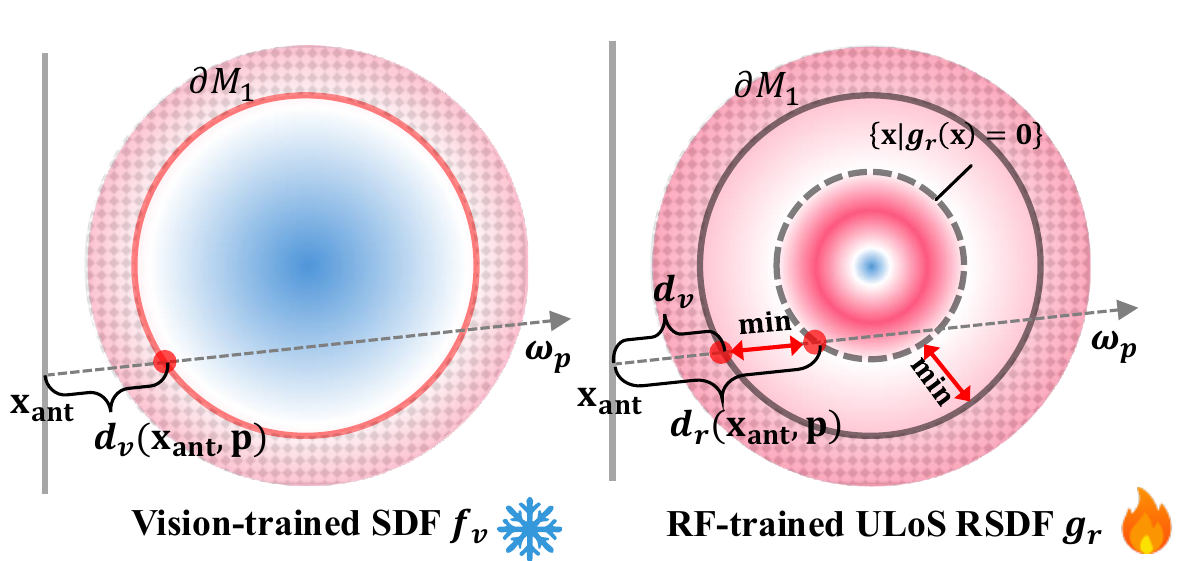}
    \caption{
Illustration of RSDF alignment.  
In the LoS region (shaded area), the vision-pretrained SDF coincides with the RF-trained ULoS RSDF.  
Therefore, the target of RSDF alignment becomes aligning the outer surface of the box,  
which can further be reduced to aligning the depth along the primary ray.
    }
    \vspace{-1.25em}
    \label{fig:rsdf}
\end{figure}

Our objective is to enforce equality between the RF-trained RSDF and the vision-trained SDF 
on a selected closed surface $S_{\mathrm{LoS}} \subset R_{0,1}$, such that
\[
    g_r(\mathbf{x}) = f_v(\mathbf{x}), 
    \qquad \forall\, \mathbf{x} \in S_{\mathrm{LoS}}.
\]
As shown in Fig.~\ref{fig:rsdf}, for implementation convenience, we choose $S_{\mathrm{LoS}} = \partial M_1$, 
that is, the outer surface of the box, as the reference surface for alignment.
The corresponding optimization objective is defined as
{\setlength{\abovedisplayskip}{5pt}%
 \setlength{\belowdisplayskip}{5pt}%
\begin{equation}
    \mathcal{L}_{\mathrm{RSDF}} 
    =
    \min_{g_r}
    \mathbb{E}_{\mathbf{x}\in \partial M_1}
    \big[
        \big| g_r(\mathbf{x}) 
        - f_v(\mathbf{x}) \big|
    \big].
\label{eq:rsdf_alignment}
\end{equation}}

However, in practice, directly sampling on the surface $\partial M_1$ 
and supervising the SDF values can be difficult and may lead to unstable training. 
Since the alignment only requires consistency of the reconstructed surfaces, 
a more stable alternative is to supervise the \emph{depth} along primary rays.

\noindent\textbf{Primary Ray Depth.}
Let $\mathbf{x}_{\mathrm{ant}}$ denote the antenna position and $\omega_p$
the ray direction. The expected depth along the primary ray for vision and RF modalities is expressed as
\begin{equation}
    d(\mathbf{x}_{\mathrm{ant}}, \omega_p) 
    = \int_0^{\infty} u\, \rho(u)\, T(u)\, du
\end{equation}
where $\rho_v$ and $T_v$ are derived from the vision-trained SDF $f_v$, 
and $\rho_r$ and $T_r$ are derived from the RF-trained RSDF $g_r$.
The final RSDF alignment loss is defined as
\begin{equation}
    \mathcal{L}_{\mathrm{RSDF}} 
    =
    \mathbb{E}_{\mathbf{x}_{\mathrm{ant}}}
    \big[
        \big| 
            d_v(\mathbf{x}_{\mathrm{ant}}, \omega_p)
            - 
            d_r(\mathbf{x}_{\mathrm{ant}}, \omega_p)
        \big|
    \big].
\label{eq:depth_alignment}
\end{equation}

\subsection{Optimization Target}

Directly training with RSDF alignment can make it difficult to reconstruct the NLoS geometry due to optimization instability.  
To address this, we disentangle the optimization process into two stages.
In \textbf{Stage 1}, as illustrated in Fig.~\ref{fig:pipeline}, we freeze the Reflectivity Network and initialize its output to a constant $1.0$ across all spatial positions. 
We adopt the training objectives proposed in \prename{}~\cite{lu2025geraf}, employing a Matched Filter (MF) on the predicted signals. 
MF effectively suppresses noise and irrelevant signal components while preserving the most informative features for geometric reconstruction. 
The resulting optimization objective is defined as:
\begin{equation}
    \mathcal{L} = \mathcal{L}_{\mathrm{MF}} + \lambda_{\mathrm{GRAD}} \mathcal{L}_{\mathrm{GRAD}},
\end{equation}
where $\mathcal{L}_{\mathrm{MF}}$ computes the loss between the predicted and ground-truth matched-filter power distributions, and $\mathcal{L}_{\mathrm{GRAD}}$ denotes the Eikonal regularization term~\cite{gropp2020implicit} used to enforce a valid signed distance field.

In {\bf Stage 2}, we train all networks jointly and apply RSDF alignment.  
Thanks to the stable initialization from Stage 1, there is no need to use ULoS Lensless Rendering in this stage.  
The overall training objective combines the matched-filter loss, RSDF alignment loss, and gradient regularization:{\setlength{\abovedisplayskip}{6pt}%
 \setlength{\belowdisplayskip}{6pt}%
\begin{equation}
    \mathcal{L} 
    = 
    \mathcal{L}_{\mathrm{MF}} 
    + 
    \lambda_{\mathrm{GRAD}}\, \mathcal{L}_{\mathrm{GRAD}}
    + 
    \lambda_{\mathrm{RSDF}}\, \mathcal{L}_{\mathrm{RSDF}}.
\end{equation}}
The RSDF alignment loss \( \mathcal{L}_{\mathrm{RSDF}} \) is defined in Eq.~\eqref{eq:depth_alignment}.

\begin{figure*}[htb]
    \centering
    \includegraphics[width=\linewidth]{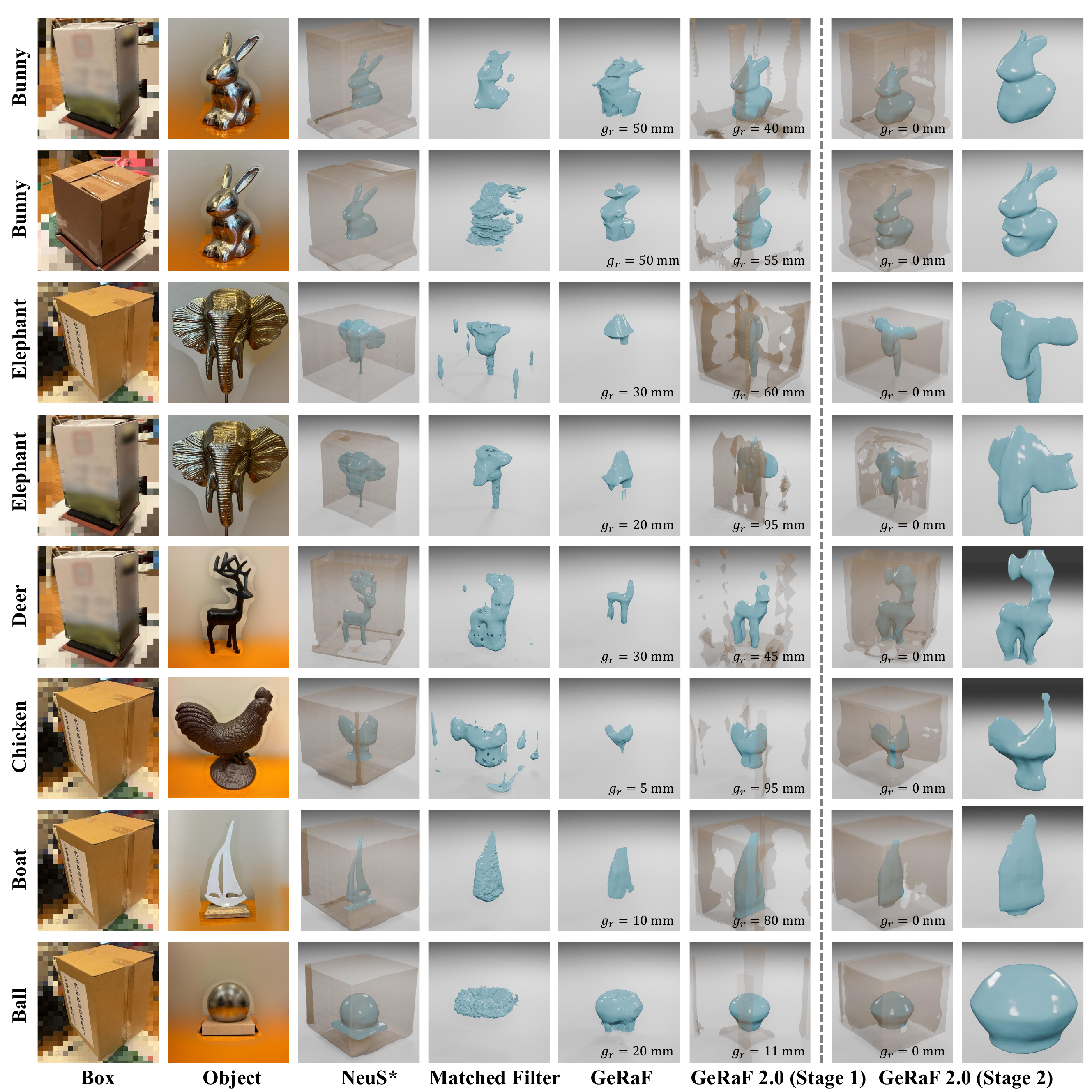}
    \caption{
Qualitative results between vision-based NeuS~\cite{wang2021neus}, point cloud-based reconstruction using matched filter, non-line-of-sight reconstruction with GeRaF~\cite{lu2025geraf}, and \name{} in Stage 1 and Stage 2.  
$^*$For NeuS, the object and box are captured separately visualized together via post-processing.
For GeRaF and Stage 1, the surface level $g_r$ is manually selected (indicated in the visualization).
}
\vspace{-1.5em}
    \label{fig:comp_box}
\end{figure*}
\section{Experiments}
\subsection{Experiment Setup}
\label{sec:imp}
\noindent\textbf{Dataset} We train and evaluate our system on a multi-view radar and camera image dataset. Data was captured with a Franka Research 3 equipped with TI's AWR1843BOOST evaluation board~\cite{ti1843}.
The object was placed on a 360\(\degree\) rotation plate, with 10\(\degree\) of rotation between each radar image. Ground truth comparison of the NLoS object for quantitative results was collected with Scaniverse~\cite{scaniverse2025}.

\noindent\textbf{Experiment Setup}
The antenna arrays are emulated using synthetic aperture radar (SAR), achieved by moving a radar sensor mounted on a robotic arm across multiple 2D planes around the object.
We simulate 36 distinct scanning poses, covering angles from $0^\circ$ to $350^\circ$ in $10^\circ$ increments.
Each scan covers an area of $0.14\,\mathrm{m} \times 0.25\,\mathrm{m}$ with antenna spacing of approximately $\frac{\lambda}{4}$.
The object is positioned approximately $0.3\,\mathrm{m}$ away from the radar.
The radar operates with a total chirp bandwidth of approximately $4\,\mathrm{GHz}$.

\noindent\textbf{Training Details}
For vision, we follow all training settings from NeuS~\cite{wang2021neus}, including model parameters, resolution, and training strategy.
For radio frequency, the SDF Network is implemented as an MLP with 8 layers and a hidden dimension of 256.
We apply sinusoidal positional encoding with 10 frequency levels as input.
The Reflectivity Network is implemented as an MLP with 4 layers and a hidden dimension of 256.
Signal power prediction is implemented as a single trainable scalar parameter.
We train our model for 100{,}000 iterations per stage (Stage 1 and Stage 2) over 48 hours on an NVIDIA H100 GPU, using \texttt{mmDetection3D}~\cite{mmdet3d2020} as the codebase.
In Stage 1, we freeze the Reflectivity Network and initialize it to output a constant value of 1.0. In Stage 2, all parameters are trained jointly.
We use an initial learning rate of \(1 \times 10^{-3}\); however, due to the sparsity of the input, the learning rate for the SDF Network is reduced to \(1 \times 10^{-4}\). Training is performed using the AdamW optimizer with cosine annealing learning rate scheduling.

\noindent\textbf{Metric Calculation}
We used the Chamfer distance and the F1-score (evaluated with a threshold of $\tau = 0.015$) for comparing our system to GeRaF and the Matched Filter. Both metrics evaluate how similar two 3D point clouds are. For mesh-to-point-cloud conversion, we uniformly sample 10{,}000 points from each mesh corresponding to the three candidate methods. We then rigidly align the matched-filter point clouds to the camera baseline, and align our outputs point clouds to the same reference.
The Chamfer distance is computed by, for each point in one cloud, finding its nearest neighbor in the other cloud and computing the squared distance. The final Chamfer distance is obtained by averaging these distances in both directions and summing the results.
The F1-score is computed by finding, for every point in one point cloud, the nearest point in the other cloud, and repeating this process in reverse. Precision (P) and Recall (R) are defined as the fractions of nearest-neighbor distances that fall below $\tau$. The F1-score is then given by
$F_1 = 2 \cdot (\text{P} \cdot \text{R})/(\text{P} + \text{R})$.

\vspace{-0.5em}
\subsection{Results}
\begin{figure}[htb]
    \centering
    \includegraphics[width=\linewidth]{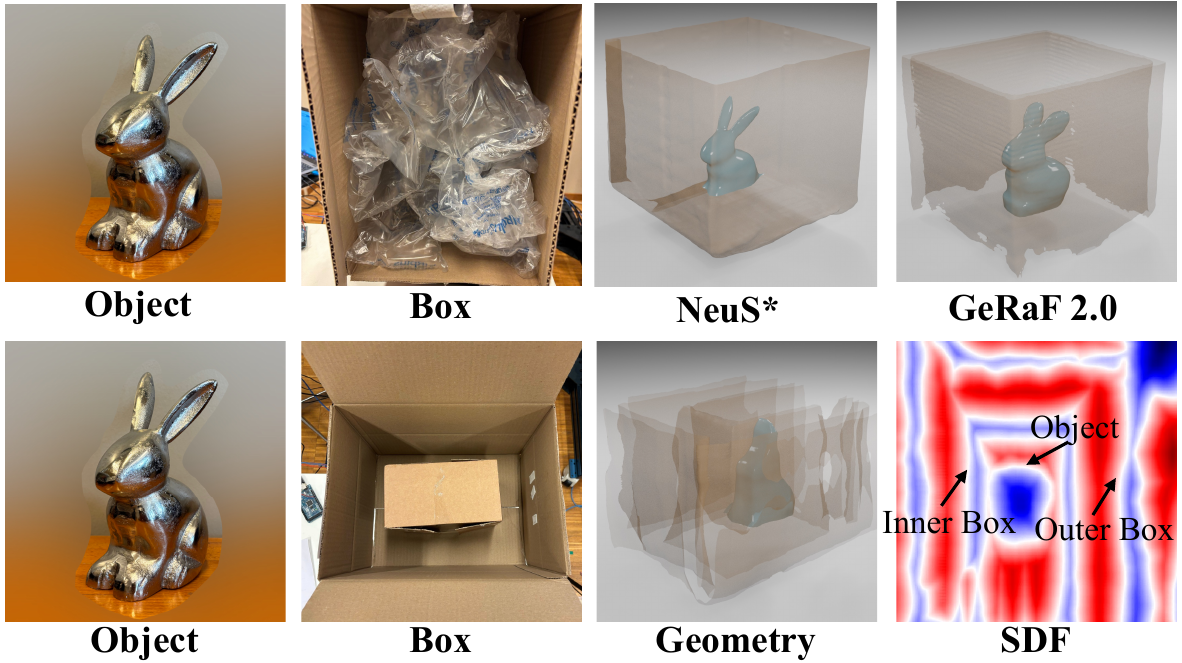}
   \caption{{\bf\em Top}: Reconstruction results for a box filled with bubble wrap.$^*$For NeuS, the object and box are captured separately visualized together via post-processing. {\bf\em Bottom}: Reconstruction of the nested box object. \texttt{Geometry} denotes \name{}'s reconstructed 3D mesh, while \texttt{SDF} illustrates the 2D SDF values for a horizontal slice.}
    \label{fig:fillnest}
\end{figure}
\begin{table}[t]
\centering
\setlength{\tabcolsep}{3pt}
\renewcommand{\arraystretch}{1.2}
\caption{Quantitative results (*different box).}
\vspace{-0.5em}
\begin{tabular}{l|ccc|ccc}
\toprule
\multirow{2}{*}{\textbf{Object}} &
\multicolumn{3}{c|}{\textbf{F1}$\uparrow$} &
\multicolumn{3}{c}{\textbf{CD (mm)}$\downarrow$} \\
\cmidrule(lr){2-4} \cmidrule(lr){5-7}
 & MF & GeRaF & Ours & MF & GeRaF & Ours \\
\midrule  
Bunny     & 0.329 & 0.822 &  \textbf{0.962} & 4.24 &  0.28 & \textbf{0.15} \\ 
Bunny*    & 0.790 & 0.859 &  \textbf{0.964} & 2.87 &  0.28 & \textbf{0.12} \\
Elephant  & 0.517 & 0.77 &  \textbf{0.845} & 9.9  &  0.41 & \textbf{0.24} \\ 
Elephant* & 0.607 & 0.568 &  \textbf{0.734} & 1.69 &  1.11 & \textbf{0.47} \\
Deer      & 0.550 & 0.643 &  \textbf{0.786} & 1.20 &  0.45 & \textbf{0.24} \\ 
Chicken   & 0.376 & 0.869 &  \textbf{0.941} & 6.19 &  0.21 & \textbf{0.14} \\
Boat      & 0.595 & 0.684 &  \textbf{0.779} & 0.93 &  0.47 & \textbf{0.43} \\
Ball      & 0.389 & 0.560 &  \textbf{0.753} & 1.42 &  0.94 & \textbf{0.52} \\
\bottomrule
\end{tabular}
\label{tab:results-quant}
\end{table}
\begin{figure}[tb]
    \centering
    \includegraphics[width=\linewidth]{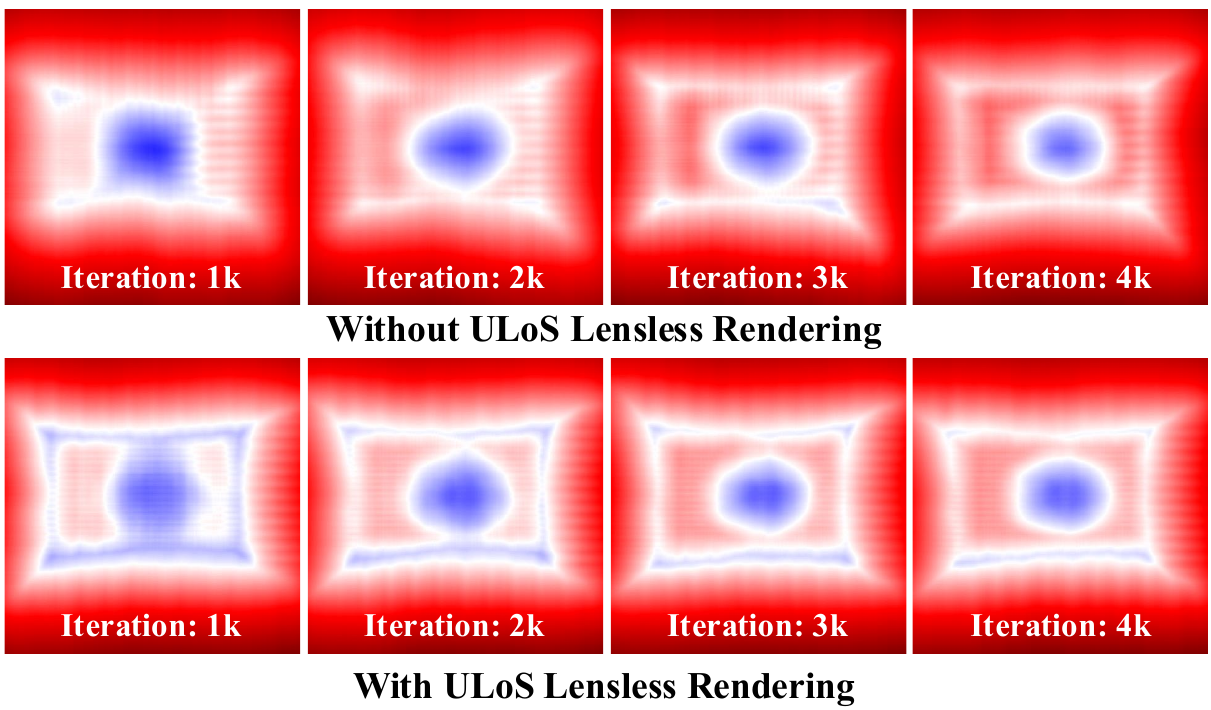}
\caption{
Ablation study on ULoS Lensless Rendering.  
We compare SDF slices of our Stage 1 baseline ({\bf\em bottom}) with a variant that does not leverage the vision-trained SDF ({\bf\em top}) during early training iterations on the \textit{bunny} object.  
Red indicates positive values, blue indicates negative values, and white represents the surface (zero level set).
}
\vspace{-1em}
    \label{fig:abl_ulos_rend}
\end{figure}
\begin{figure}[tb]
    \centering
    \includegraphics[width=\linewidth]{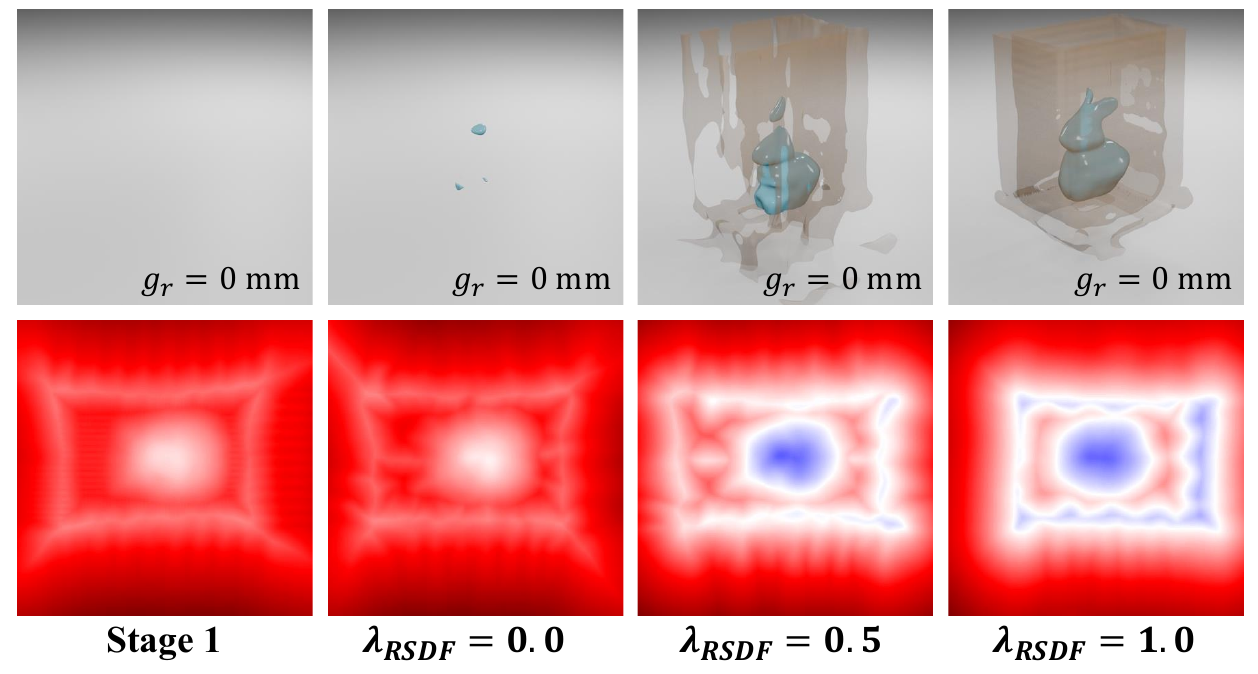}
    \caption{
Ablation study on different values of $\lambda_\mathrm{RSDF}$. 
The first column shows Stage 1 results. 
The top row visualizes surfaces at the zero-level set of the SDF, while the bottom row shows a cross-sectional slice of the SDF.
}
\vspace{-1em}
    \label{fig:abl_lrsdf}
\end{figure}
\begin{figure}[tb]
    \centering
    \includegraphics[width=\linewidth]{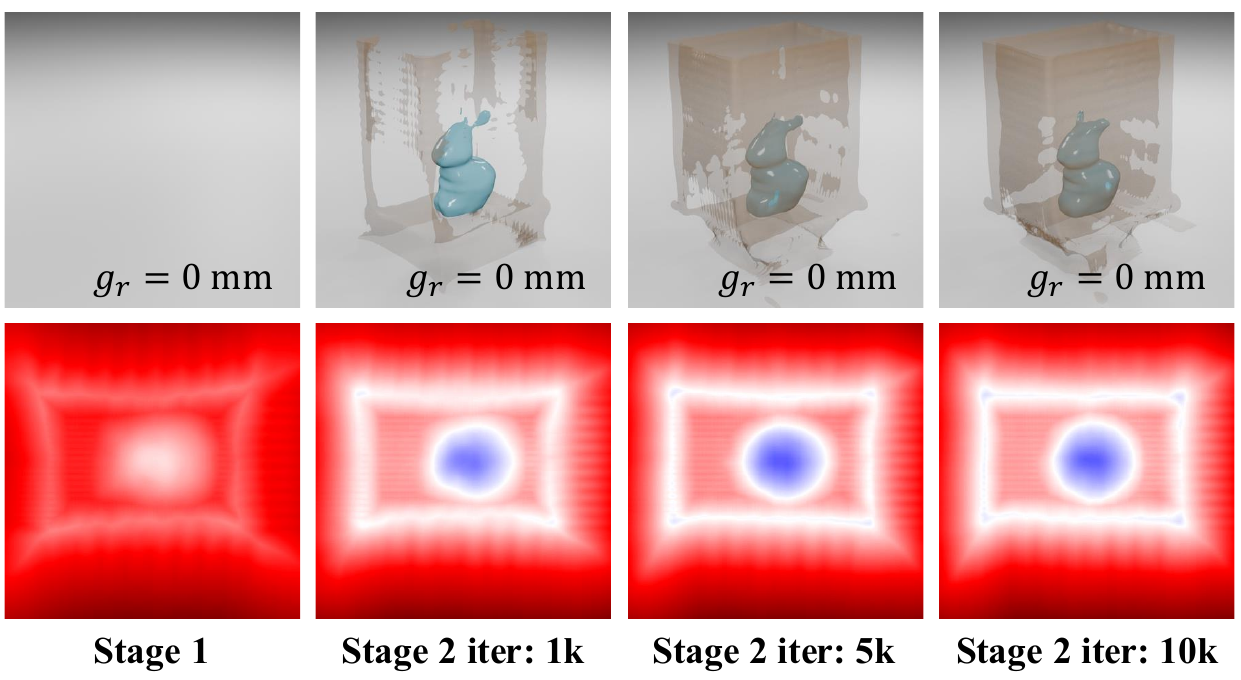}
   \caption{
Effect of RSDF alignment during training, from Stage 1 (initialization) to 10{,}000 iterations.  
We show 3D reconstruction results ({\bf\em top}) and SDF slices ({\bf\em bottom}) of the \textit{bunny} object, taken at Stage 1, and at 1{,}000, 5{,}000, and 10{,}000 iterations of Stage 2.  
Red indicates positive SDF values, blue indicates negative values, and white denotes the surface (zero level set).
}
\vspace{-1em}
    \label{fig:abl_stage2_iter}
\end{figure}
\begin{figure*}[tb]
    \centering
    \includegraphics[width=\linewidth]{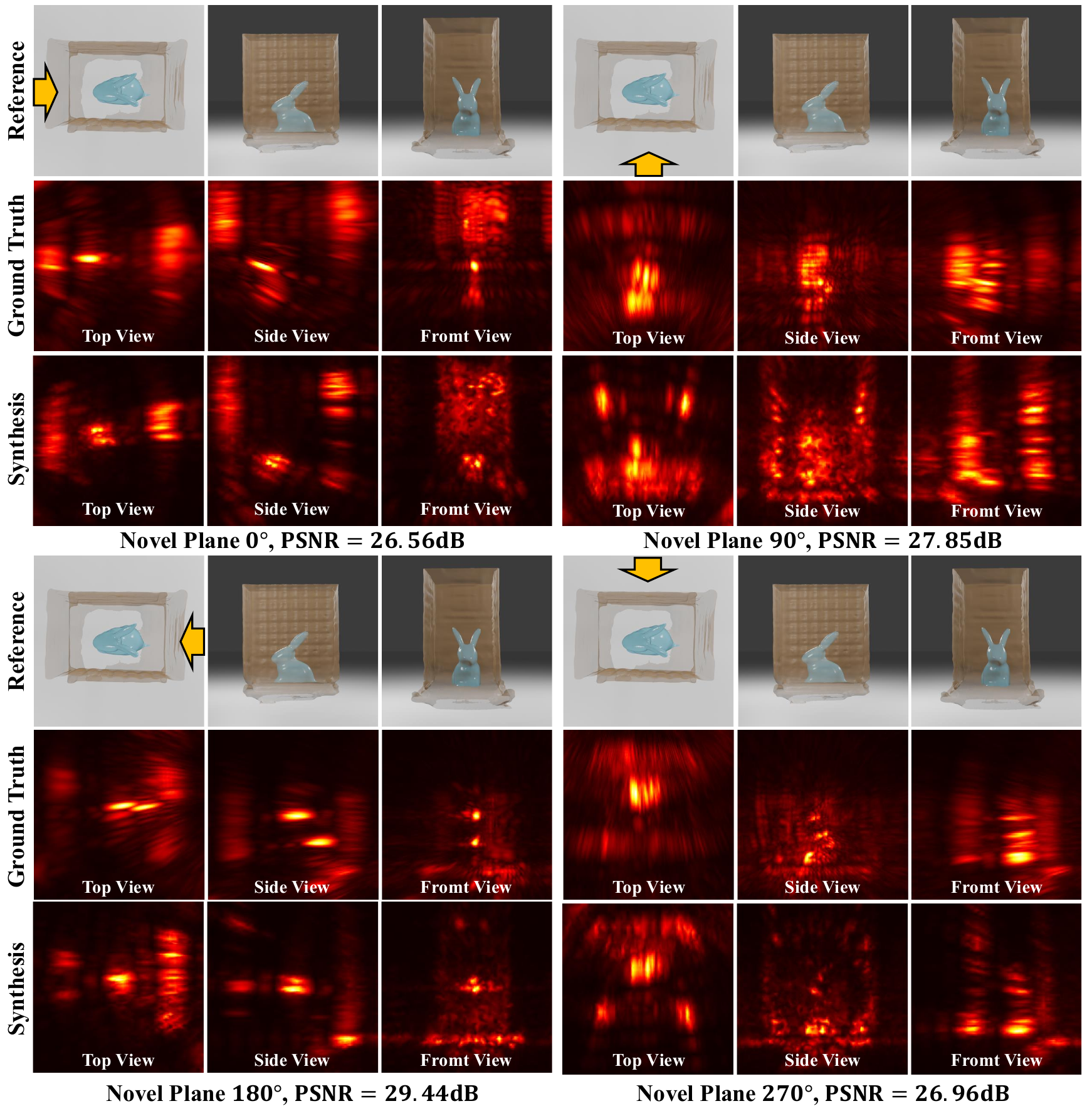}
    \caption{
Novel view synthesis on antenna planes at $0^\circ$, $90^\circ$, $180^\circ$, and $270^\circ$.  
For each view, we show the corresponding reference image from the vision modality for comparison.  
The orange arrow indicates the incoming signal direction.  
The 3D matched filter results are projected onto the three axes using maximum intensity projection.
}
    \vspace{-1.25em}
    \label{fig:nvs_bunnybv3}
\end{figure*}
\vspace{-0.25em}
\label{sec:results}
We compare \name{} with three baselines: vision-based NeuS~\cite{wang2021neus}, Matched Filter (MF) imaging, and NLoS reconstruction GeRaF~\cite{lu2025geraf}. 
For NeuS, we train the object and box separately due to occlusion. 
For the Matched Filter, we sum outputs across views, threshold the heatmap, and apply Poisson surface reconstruction. 
We show results for both Stage 1 and Stage 2 of \name{}. Quantitative results are calculated using the F1-Score(\(\tau=0.015\)) and Chamfer Distance (in millimeters), the box is cropped out of the point clouds for evaluation; the method is described in detail in the appendix.

\noindent \textbf{Qualitative Baseline Comparisons}
Qualitative results are shown in Fig.~\ref{fig:comp_box} and both stages of \name{} achieve the best RF reconstruction results, outperforming the Matched Filter and GeRaF~\cite{lu2025geraf}, and closely matching the vision-based reconstruction.
Unlike GeRaF, which often shows artifacts or fails to capture surfaces correctly, \name{} produces clean and accurate geometry due to the ULoS representation and ULoS rendering.
Notably, Stage 2 benefits from RSDF alignment, enabling the surface to be extracted precisely at the SDF zero-level, something all baselines fail to do.
This not only removes the need for manual thresholding but preserves fine details such as the elephant’s tusks, the chicken’s comb, the deer’s antlers, and the ball’s top.

\noindent \textbf{Quantitative Baseline Comparisons}
Quantitative results are shown in Tab.~\ref{tab:results-quant}. \name{} shows clear improvements in both F1-score and Chamfer distance (CD) over both the matched filter and GeRaF. It is notable that when the Matched Filter result has a high F1-score or CD both GeRaF and \name{} report higher scores.

\noindent \textbf{Robustness to Fillers and Multilayer Occlusions.}
The first row of Fig.~\ref{fig:fillnest} illustrates the reconstruction of the target object when the container is filled with bubble wrap. Despite the introduction of additional scattering layers, the model maintains high-fidelity reconstruction with negligible degradation in geometric detail.
Furthermore, the second row of Fig.~\ref{fig:fillnest} evaluates the system's performance in a nested-box configuration. The corresponding SDF visualization clearly resolves the zero-level sets for both the concentric boxes and the internal object albeit some degradation of the internal surface due to increased scattering.

\noindent \textbf{Novel View Synthesis}
To evaluate novel view synthesis (NVS), we remove antenna planes at angles $0^\circ$, $90^\circ$, $180^\circ$, and $270^\circ$ from the training set and include them in the evaluation set.
Fig.~\ref{fig:nvs_bunnybv3} shows qualitative and quantitative (PSNR~\cite{mildenhall2021nerf}) NVS results for the \textit{bunny} object.
Unlike NVS in the visible light spectrum, mmWave synthesis typically lacks high-frequency spatial features. However, the reconstruction of low-frequency structures and the relatively high PSNR values demonstrate the model's potential capability to learn both the underlying geometry and the specific characteristics of mmWave imaging.

\subsection{Ablation Studies}

\noindent\textbf{ULoS Lensless Rendering}
In Fig.~\ref{fig:abl_ulos_rend}, we compare the effect of ULoS Lensless Rendering.
This module leverages the vision-trained SDF as guidance to adjust the accumulated transmittance.
We visualize a slice of the SDF during the early stage of Stage 1 training on \textit{bunny} to observe the impact on convergence.
With this guidance, the results shown at the bottom of Fig.~\ref{fig:abl_ulos_rend} demonstrate faster and more accurate shape formation for both the box and the object.

\noindent\textbf{RSDF Alignment}
In Fig.~\ref{fig:abl_lrsdf}, we show the impact of RSDF alignment during Stage 2.
All visualizations display surfaces extracted at the zero level set of the SDF.
Without alignment (i.e., $\lambda_\mathrm{RSDF}=0$), the reconstructed surface is significantly offset from the zero level.
As we increase $\lambda_\mathrm{RSDF}$, the surface gradually converges toward the correct zero level, demonstrating the effectiveness of RSDF alignment in resolving surface ambiguity.

\noindent\textbf{Additional RSDF Alignment Studies}
We include additional ablation studies.
In Fig.~\ref{fig:abl_stage2_iter}, we visualize the reconstruction results and SDF slices during Stage 2 training on the \textit{bunny} inside the box.
With the help of primary depth supervision from the vision-trained SDF, RSDF alignment quickly brings the zero level set to the correct surface within only 10{,}000 iterations.

\section{Conclusion}
\label{sec:conc}
In this paper, we present a unified neural reconstruction framework that bridges LoS and NLoS regions for 3D reconstruction.
By incorporating LoS geometry,
as a physical prior into the neural field formulation, our method establishes a consistent link between visible and hidden surfaces, stabilizing optimization and improving reconstruction.

\section*{Acknowledgments}
\vspace{-0.5em}
We thank the anonymous CVPR reviewers and members of the SENS Lab for their valuable feedback. We would also like to thank Ralf Boehnke, Dymtro Rachkov and Daniel Ardila Palomino from Sony Research for their helpful feedback. This project is funded in part by the Sony Faculty Innovation Fellowship.

{
    \small
    \bibliographystyle{ieeenat_fullname}
    \bibliography{references}

@STRING{cvpr	= {CVPR} }

@STRING{eccv	= {ECCV} }

@STRING{iccv	= {ICCV} }

@STRING{nips	= {NeurIPS} }

@STRING{iclr	= {ICLR} }

@STRING{sigcomm	= {SIGCOMM} }

@STRING{mobicom	= {MobiCom} }

@STRING{siggraph = {SIGGRAPH}}

@STRING{tog	= {ACM Transactions on Graphics (TOG)} }

@inproceedings{wang2021neus,
  title={NeuS: Learning Neural Implicit Surfaces by Volume Rendering for Multi-view Reconstruction},
  author={Wang, Peng and Liu, Lingjie and Liu, Yuan and Theobalt, Christian and Komura, Taku and Wang, Wenping},
  booktitle=nips,
  year={2021}
}

@article{muller2022instant,
  title={Instant neural graphics primitives with a multiresolution hash encoding},
  author={M{\"u}ller, Thomas and Evans, Alex and Schied, Christoph and Keller, Alexander},
  journal=tog,
  year={2022},
}

@inproceedings{munkberg2022extracting,
  title={Extracting triangular 3d models, materials, and lighting from images},
  author={Munkberg, Jacob and Hasselgren, Jon and Shen, Tianchang and Gao, Jun and Chen, Wenzheng and Evans, Alex and M{\"u}ller, Thomas and Fidler, Sanja},
  booktitle=cvpr,
  year={2022}
}

@article{liu2023nero,
  title={Nero: Neural geometry and brdf reconstruction of reflective objects from multiview images},
  author={Liu, Yuan and Wang, Peng and Lin, Cheng and Long, Xiaoxiao and Wang, Jiepeng and Liu, Lingjie and Komura, Taku and Wang, Wenping},
  journal=tog,
  year={2023},
}

@inproceedings{guan2020through,
  title={Through fog high-resolution imaging using millimeter wave radar},
  author={Guan, Junfeng and Madani, Sohrab and Jog, Suraj and Gupta, Saurabh and Hassanieh, Haitham},
  booktitle=cvpr,
  year={2020}
}

@inproceedings{adib2013see,
  title={See through walls with WiFi!},
  author={Adib, Fadel and Katabi, Dina},
  booktitle=sigcomm,
  year={2013}
}

@inproceedings{yanik2019near,
  title={Near-field 2-D SAR imaging by millimeter-wave radar for concealed item detection},
  author={Yanik, Muhammet Emin and Torlak, Murat},
  booktitle={2019 IEEE radio and Wireless Symposium (RWS)},
  year={2019},
}

@article{takawale2025spinr,
  title={SpINR: Neural Volumetric Reconstruction for FMCW Radars},
  author={Takawale, Harshvardhan and Roy, Nirupam},
  journal={arXiv preprint},
  year={2025}
}

@inproceedings{niemeyer2020differentiable,
  title={Differentiable volumetric rendering: Learning implicit 3d representations without 3d supervision},
  author={Niemeyer, Michael and Mescheder, Lars and Oechsle, Michael and Geiger, Andreas},
  booktitle=cvpr,
  year={2020}
}

@inproceedings{yariv2021volume,
  title={Volume rendering of neural implicit surfaces},
  author={Yariv, Lior and Gu, Jiatao and Kasten, Yoni and Lipman, Yaron},
  booktitle=nips,
  year={2021}
}

@inproceedings{yariv2020multiview,
  title={Multiview neural surface reconstruction by disentangling geometry and appearance},
  author={Yariv, Lior and Kasten, Yoni and Moran, Dror and Galun, Meirav and Atzmon, Matan and Ronen, Basri and Lipman, Yaron},
  booktitle=nips,
  year={2020}
}

@inproceedings{huang20242d,
  title={2d gaussian splatting for geometrically accurate radiance fields},
  author={Huang, Binbin and Yu, Zehao and Chen, Anpei and Geiger, Andreas and Gao, Shenghua},
  booktitle=siggraph,
  year={2024}
}

@inproceedings{borts2024radar,
  title={Radar fields: Frequency-space neural scene representations for fmcw radar},
  author={Borts, David and Liang, Erich and Broedermann, Tim and Ramazzina, Andrea and Walz, Stefanie and Palladin, Edoardo and Sun, Jipeng and Brueggemann, David and Sakaridis, Christos and Van Gool, Luc and others},
  booktitle=siggraph,
  year={2024}
}

@inproceedings{boss2021nerd,
  title={Nerd: Neural reflectance decomposition from image collections},
  author={Boss, Mark and Braun, Raphael and Jampani, Varun and Barron, Jonathan T and Liu, Ce and Lensch, Hendrik},
  booktitle=iccv,
  year={2021}
}

@article{chen2023neusg,
  title={Neusg: Neural implicit surface reconstruction with 3d gaussian splatting guidance},
  author={Chen, Hanlin and Li, Chen and Lee, Gim Hee},
  journal={arXiv preprint},
  year={2023}
}

@inproceedings{gao2024relightable,
  title={Relightable 3d gaussians: Realistic point cloud relighting with brdf decomposition and ray tracing},
  author={Gao, Jian and Gu, Chun and Lin, Youtian and Li, Zhihao and Zhu, Hao and Cao, Xun and Zhang, Li and Yao, Yao},
  booktitle=eccv,
  year={2024},
}

@inproceedings{huang2024dart,
  title={DART: Implicit doppler tomography for radar novel view synthesis},
  author={Huang, Tianshu and Miller, John and Prabhakara, Akarsh and Jin, Tao and Laroia, Tarana and Kolter, Zico and Rowe, Anthony},
  booktitle=cvpr,
  year={2024}
}

@inproceedings{jiang2024gaussianshader,
  title={Gaussianshader: 3d gaussian splatting with shading functions for reflective surfaces},
  author={Jiang, Yingwenqi and Tu, Jiadong and Liu, Yuan and Gao, Xifeng and Long, Xiaoxiao and Wang, Wenping and Ma, Yuexin},
  booktitle=cvpr,
  year={2024}
}

@inproceedings{jin2023tensoir,
  title={Tensoir: Tensorial inverse rendering},
  author={Jin, Haian and Liu, Isabella and Xu, Peijia and Zhang, Xiaoshuai and Han, Songfang and Bi, Sai and Zhou, Xiaowei and Xu, Zexiang and Su, Hao},
  booktitle=cvpr,
  year={2023}
}

@article{kerbl20233d,
  title={3d gaussian splatting for real-time radiance field rendering.},
  author={Kerbl, Bernhard and Kopanas, Georgios and Leimk{\"u}hler, Thomas and Drettakis, George},
  journal=tog,
  year={2023}
}

@inproceedings{liang2024gs,
  title={Gs-ir: 3d gaussian splatting for inverse rendering},
  author={Liang, Zhihao and Zhang, Qi and Feng, Ying and Shan, Ying and Jia, Kui},
  booktitle=cvpr,
  year={2024}
}

@article{mildenhall2021nerf,
  title={Nerf: Representing scenes as neural radiance fields for view synthesis},
  author={Mildenhall, Ben and Srinivasan, Pratul P and Tancik, Matthew and Barron, Jonathan T and Ramamoorthi, Ravi and Ng, Ren},
  journal={Communications of the ACM},
  year={2021},
}

@article{richards2010principles,
  title={Principles of modern radar},
  author={Richards, Mark A and Scheer, Jim and Holm, William A and Melvin, William L},
  year={2010},
  publisher={Citeseer}
}

@inproceedings{srinivasan2021nerv,
  title={Nerv: Neural reflectance and visibility fields for relighting and view synthesis},
  author={Srinivasan, Pratul P and Deng, Boyang and Zhang, Xiuming and Tancik, Matthew and Mildenhall, Ben and Barron, Jonathan T},
  booktitle=cvpr,
  year={2021}
}

@inproceedings{sun20213drimr,
  title={3DRIMR: 3D reconstruction and imaging via mmWave radar based on deep learning},
  author={Sun, Yue and Huang, Zhuoming and Zhang, Honggang and Cao, Zhi and Xu, Deqiang},
  booktitle={IEEE International Performance, Computing, and Communications Conference (IPCCC)},
  year={2021},
}

@inproceedings{tan2024fast,
  title={Fast 3D Reconstruction of Space Targets From ISAR Image Sequences based on Neural Network},
  author={Tan, Weiyi and Wang, Yu and Tian, Biao and Xu, Shiyou},
  booktitle={IEEE 8th International Conference on Vision, Image and Signal Processing (ICVISP)},
  year={2024},
}

@inproceedings{yao2022neilf,
  title={Neilf: Neural incident light field for physically-based material estimation},
  author={Yao, Yao and Zhang, Jingyang and Liu, Jingbo and Qu, Yihang and Fang, Tian and McKinnon, David and Tsin, Yanghai and Quan, Long},
  booktitle=eccv,
  year={2022},
}

@article{wu2015safe,
  title={Safe for generations to come: Considerations of safety for millimeter waves in wireless communications},
  author={Wu, Ting and Rappaport, Theodore S and Collins, Christopher M},
  journal={IEEE microwave magazine},
  year={2015}
}

@inproceedings{lai2024enabling,
  title={Enabling Visual Recognition at Radio Frequency},
  author={Lai, Haowen and Luo, Gaoxiang and Liu, Yifei and Zhao, Mingmin},
  booktitle=mobicom,
  year={2024}
}

@inproceedings{sun20223d,
  title={3D reconstruction of multiple objects by mmWave radar on UAV},
  author={Sun, Yue and Huang, Zhuoming and Zhang, Honggang and Liang, Xiaohui},
  booktitle={2022 IEEE 19th International Conference on Mobile Ad Hoc and Smart Systems (MASS)},
  year={2022}
}

@inproceedings{sun2022r2p,
  title={R2p: A deep learning model from mmwave radar to point cloud},
  author={Sun, Yue and Zhang, Honggang and Huang, Zhuoming and Liu, Benyuan},
  booktitle={International Conference on Artificial Neural Networks},
  year={2022}
}

@misc{barbierrenard2025multiview3dsurfacereconstruction,
      title={Multi-view 3D surface reconstruction from SAR images by inverse rendering}, 
      author={Emile Barbier--Renard and Florence Tupin and Nicolas Trouvé and Loïc Denis},
      year={2025},
      archivePrefix={arXiv},
}

@inproceedings{gu2025irgs,
  title={Irgs: Inter-reflective gaussian splatting with 2d gaussian ray tracing},
  author={Gu, Chun and Wei, Xiaofei and Zeng, Zixuan and Yao, Yuxuan and Zhang, Li},
  booktitle=cvpr,
  year={2025}
}

@inproceedings{yao2025reflective,
  title={Reflective Gaussian Splatting},
  author={Yao, Yuxuan and Zeng, Zixuan and Gu, Chun and Zhu, Xiatian and Zhang, Li},
  booktitle=iclr,
  year={2025}
}

@misc{ti1843,
	author =       {{TI Inc.}},
	title =        {Texas Instrument AWR1843},
	howpublished = {\url{https://www.ti.com/product/AWR1843}},
	year =        {2023}
}

@article{andersson2012fast,
  title={Fast Fourier methods for synthetic aperture radar imaging},
  author={Andersson, Fredrik and Moses, Randolph and Natterer, Frank},
  journal={IEEE Transactions on Aerospace and Electronic Systems},
  year={2012}
}

@misc{mmdet3d2020,
    title={{MMDetection3D: OpenMMLab} next-generation platform for general {3D} object detection},
    author={MMDetection3D Contributors},
    howpublished = {\url{https://github.com/open-mmlab/mmdetection3d}},
    year={2020}
}

@inproceedings{lu2025geraf,
  title={GeRaF: Neural Geometry Reconstruction from Radio Frequency Signals},
  author={Lu, Jiachen and Shanbhag, Hailan and Hassanieh, Haitham},
  booktitle=nips,
  year={2025},
}

@inproceedings{dodds2025non,
  title={Non-Line-of-Sight 3D Object Reconstruction via mmWave Surface Normal Estimation},
  author={Dodds, Laura and Boroushaki, Tara and Zhou, Kaichen and Adib, Fadel},
  booktitle=mobicom,
  year={2025}
}

@article{hussein20253d,
  title={3D Object Reconstruction with mmWave Radars},
  author={Hussein, Samah and Guan, Junfeng and Narashiman, Swathi and Gupta, Saurabh and Hassanieh, Haitham},
  journal={arXiv preprint arXiv:2504.12348},
  year={2025}
}

@article{zhang2025rf4d,
  title={RF4D: Neural Radar Fields for Novel View Synthesis in Outdoor Dynamic Scenes},
  author={Zhang, Jiarui and Li, Zhihao and Wang, Chong and Wen, Bihan},
  journal={arXiv preprint arXiv:2505.20967},
  year={2025}
}

@inproceedings{kung2025radarsplat,
  title={Radarsplat: Radar gaussian splatting for high-fidelity data synthesis and 3d reconstruction of autonomous driving scenes},
  author={Kung, Pou-Chun and Harisha, Skanda and Vasudevan, Ram and Eid, Aline and Skinner, Katherine A},
  booktitle={Proceedings of the IEEE/CVF International Conference on Computer Vision},
  pages={27596--27606},
  year={2025}
}

@article{wu2023mvfusion,
  title={Mvfusion: Multi-view 3d object detection with semantic-aligned radar and camera fusion},
  author={Wu, Zizhang and Chen, Guilian and Gan, Yuanzhu and Wang, Lei and Pu, Jian},
  journal={arXiv preprint arXiv:2302.10511},
  year={2023}
}

@article{el2015toward,
  title={Toward 3D reconstruction of outdoor scenes using an MMW radar and a monocular vision sensor},
  author={El Natour, Ghina and Ait-Aider, Omar and Rouveure, Raphael and Berry, Fran{\c{c}}ois and Faure, Patrice},
  journal={Sensors},
  volume={15},
  number={10},
  pages={25937--25967},
  year={2015},
  publisher={MDPI}
}

@inproceedings{lin2024rcbevdet,
  title={Rcbevdet: Radar-camera fusion in bird's eye view for 3d object detection},
  author={Lin, Zhiwei and Liu, Zhe and Xia, Zhongyu and Wang, Xinhao and Wang, Yongtao and Qi, Shengxiang and Dong, Yang and Dong, Nan and Zhang, Le and Zhu, Ce},
  booktitle={Proceedings of the IEEE/CVF Conference on Computer Vision and Pattern Recognition},
  pages={14928--14937},
  year={2024}
}

@inproceedings{dhekne2018liquid,
  title={Liquid: A wireless liquid identifier},
  author={Dhekne, Ashutosh and Gowda, Mahanth and Zhao, Yixuan and Hassanieh, Haitham and Choudhury, Romit Roy},
  booktitle={Proceedings of the 16th annual international conference on mobile systems, applications, and services},
  pages={442--454},
  year={2018}
}

@article{chen2024towards,
  title={Towards weather-robust 3D human body reconstruction: Millimeter-wave radar-based dataset, benchmark, and multi-modal fusion},
  author={Chen, Anjun and Wang, Xiangyu and Shi, Kun and Huo, Yuchi and Chen, Jiming and Ye, Qi},
  journal={IEEE Transactions on Circuits and Systems for Video Technology},
  year={2024},
  publisher={IEEE}
}

@inproceedings{yang2025zfusion,
  title={ZFusion: An Effective Fuser of Camera and 4D Radar for 3D Object Perception in Autonomous Driving},
  author={Yang, Sheng and Zhan, Tong and Qiao, Shichen and Gong, Jicheng and Yang, Qing and Wang, Jian and Lu, Yanfeng},
  booktitle={Proceedings of the Computer Vision and Pattern Recognition Conference},
  pages={3768--3777},
  year={2025}
}

@article{ali2019multi,
  title={Multi-sensor depth fusion framework for real-time 3D reconstruction},
  author={Ali, Muhammad Kashif and Rajput, Asif and Shahzad, Muhammad and Khan, Farhan and Akhtar, Faheem and B{\"o}rner, Anko},
  journal={Ieee Access},
  volume={7},
  pages={136471--136480},
  year={2019},
  publisher={IEEE}
}

@article{yu2023sparsefusion3d,
  title={SparseFusion3D: Sparse sensor fusion for 3D object detection by radar and camera in environmental perception},
  author={Yu, Zedong and Wan, Weibing and Ren, Maiyu and Zheng, Xiuyuan and Fang, Zhijun},
  journal={IEEE Transactions on Intelligent Vehicles},
  volume={9},
  number={1},
  pages={1524--1536},
  year={2023},
  publisher={IEEE}
}

@article{stone1997electromagnetic,
  title={Electromagnetic signal attenuation in construction materials},
  author={Stone, William C},
  year={1997},
  publisher={William C. Stone}
}

@software{scaniverse2025,
  title        = {Scaniverse: 3D Scanner App},
  author       = {{Toolbox AI, Inc. and Niantic, Inc.}},
  year         = {2025},
  url          = {https://www.scaniverse.com},
  version      = {5.1.0},
  note         = {Mobile application for iOS}
}

@article{gropp2020implicit,
  title={Implicit geometric regularization for learning shapes},
  author={Gropp, Amos and Yariv, Lior and Haim, Niv and Atzmon, Matan and Lipman, Yaron},
  journal={arXiv preprint arXiv:2002.10099},
  year={2020}
}

@article{yue2020bodycompass,
  title={BodyCompass: Monitoring sleep posture with wireless signals},
  author={Yue, Shichao and Yang, Yuzhe and Wang, Hao and Rahul, Hariharan and Katabi, Dina},
  journal={Proceedings of the ACM on Interactive, Mobile, Wearable and Ubiquitous Technologies},
  volume={4},
  number={2},
  pages={1--25},
  year={2020},
  publisher={ACM New York, NY, USA}
}

@inproceedings{chenradarsim,
  title={RadarSim: Simulating Single-Chip Radar via Multimodal Neural Fields},
  author={Chen, Chuhan and Huang, Tianshu and Prabhakara, Akarsh and Mummadi, Chaithanya Kumar and Cong, Zhongxiao and Rowe, Anthony and O'Toole, Matthew and Ramanan, Deva},
  booktitle={Thirteenth International Conference on 3D Vision}
}

@article{o2018confocal,
  title={Confocal non-line-of-sight imaging based on the light-cone transform},
  author={O’Toole, Matthew and Lindell, David B and Wetzstein, Gordon},
  journal={Nature},
  volume={555},
  number={7696},
  pages={338--341},
  year={2018},
  publisher={Nature Publishing Group UK London}
}

@article{liu2019non,
  title={Non-line-of-sight imaging using phasor-field virtual wave optics},
  author={Liu, Xiaochun and Guill{\'e}n, Ib{\'o}n and La Manna, Marco and Nam, Ji Hyun and Reza, Syed Azer and Huu Le, Toan and Jarabo, Adrian and Gutierrez, Diego and Velten, Andreas},
  journal={Nature},
  volume={572},
  number={7771},
  pages={620--623},
  year={2019},
  publisher={Nature Publishing Group UK London}
}

@article{lindell2019wave,
  title={Wave-based non-line-of-sight imaging using fast fk migration},
  author={Lindell, David B and Wetzstein, Gordon and O'Toole, Matthew},
  journal={ACM Transactions on Graphics (ToG)},
  volume={38},
  number={4},
  pages={1--13},
  year={2019},
  publisher={ACM New York, NY, USA}
}

@article{velten2012recovering,
  title={Recovering three-dimensional shape around a corner using ultrafast time-of-flight imaging},
  author={Velten, Andreas and Willwacher, Thomas and Gupta, Otkrist and Veeraraghavan, Ashok and Bawendi, Moungi G and Raskar, Ramesh},
  journal={Nature communications},
  volume={3},
  number={1},
  pages={745},
  year={2012},
  publisher={Nature Publishing Group UK London}
}
}

\appendix
\section{Additional Experiments}
\label{sec:exp-supp}

\begin{figure}[htb]
\vspace{-1em}
    \centering
    \includegraphics[width=\linewidth]{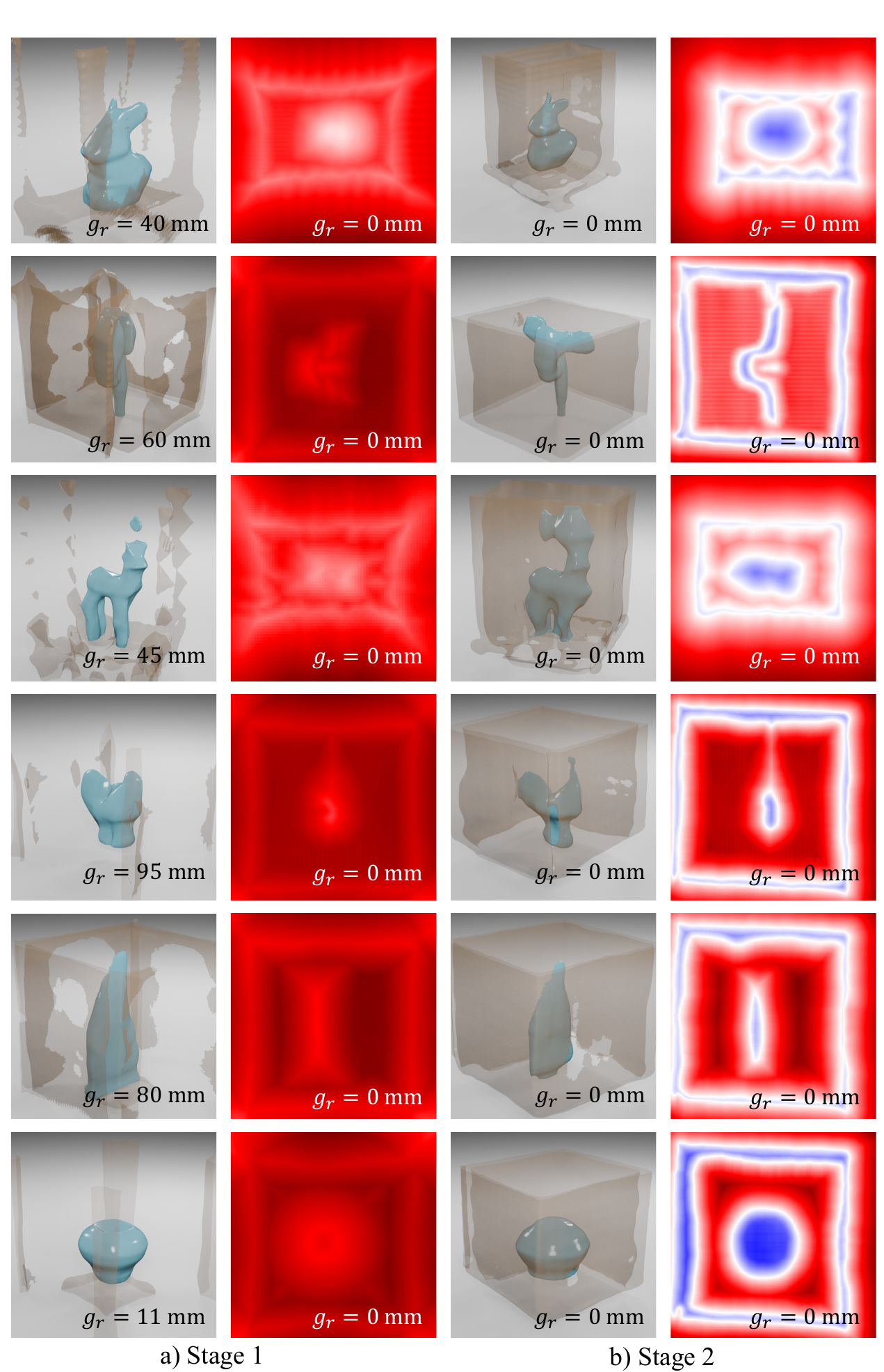}
    \caption{
Ablation study on RSDF alignment. We show both 3D reconstruction results and SDF slices.  
(a) shows the results from Stage 1, and (b) shows the results after Stage 2 training.
}
\vspace{-1em}
    \label{fig:abl_lrsdf_supp}
\end{figure}
\subsection{Additional RSDF Alignment Ablation}
In Fig.~\ref{fig:abl_lrsdf_supp}, we present additional ablation studies on RSDF alignment by comparing SDF slices for different objects.
Without RSDF alignment, the RF-trained SDF fails to reach the correct zero level set, and the reconstructed shape is also inaccurate.
In contrast, with RSDF alignment, the surface converges to the correct zero level set and the overall shape reconstruction is significantly improved.

\begin{figure}[htb]
\vspace{-1em}
    \centering
    \includegraphics[width=\linewidth]{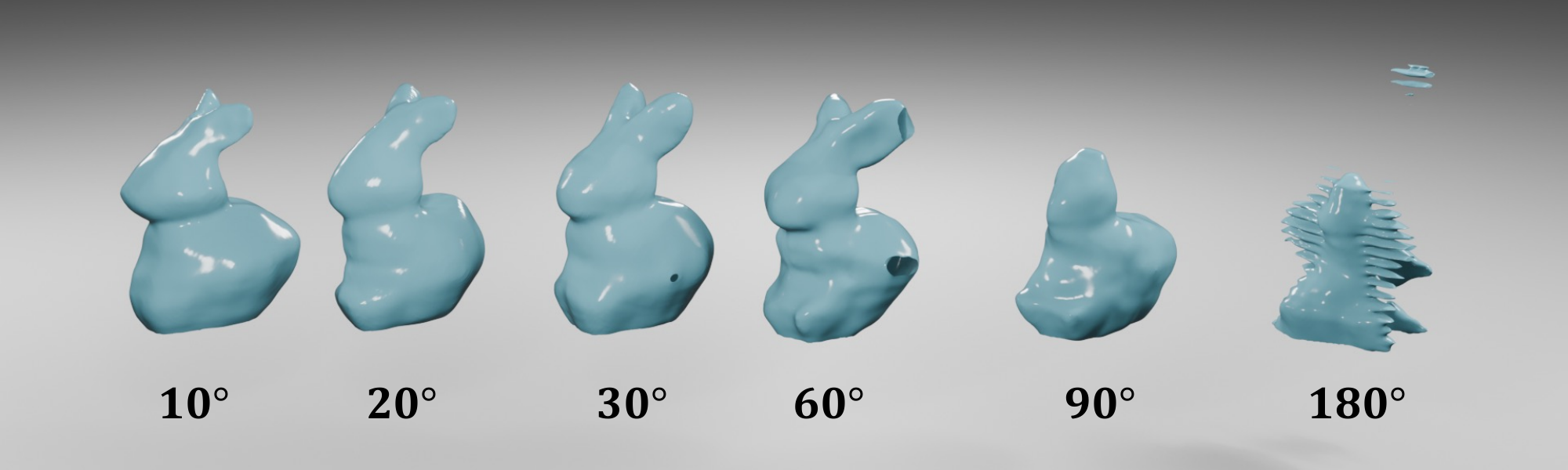}
    \caption{
Ablation study on number of scanning planes used in training.
}
\vspace{-1em}
    \label{fig:abl_scans}
\end{figure}
\subsection{Number of Scanning Plane Impact}
In the main paper, we use 36 measurements per object, with a 10$^\circ$ angular interval between viewing directions. 
We include an ablation study evaluating reconstruction quality relative to the number of scanning planes (images) used. Fig.~\ref{fig:abl_scans} shows evaluation against varying measurement densities using 20$^\circ$, 30$^\circ$, 60$^\circ$, 90$^\circ$ and 180$^\circ$ intervals. The results show that \name{} maintains strong reconstruction quality even with heavily reduced data. Using only 1/6 of the measurements (60$^\circ$) still recovers the overall geometry, though some regions (such as the bunny's back) are missing, and using 1/9 of the measurements (90$^\circ$) maintains the bunny's body, though loses the ear reconstruction. Given the minor performance drop, certain applications benefit from lower training times while preserving an acceptable reconstruction quality. 

\subsection{Non-Specular Objects}
\begin{figure}[htb]
    \centering
    \includegraphics[width=\linewidth]{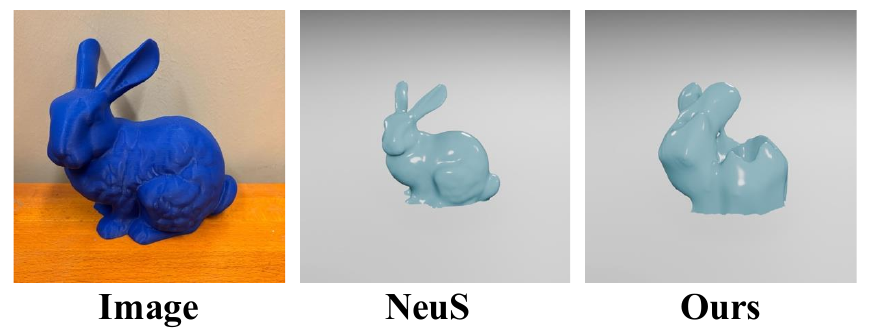}
   \caption{Reconstruction results for non-specular objects.  }
\vspace{-1em}
    \label{fig:nonmetal}
\end{figure}

While our formulation proposed uses a specular-reflection model, 
it can handle some diffusion because our ray tracer uses a small angular spread around the specular direction.
To demonstrate this, we provide an initial results for  
a 3D printed bunny (PLA) in Fig.~\ref{fig:nonmetal},
showing consistent performance for this material type. 
 While this model is sufficient for the objects and materials tested, it doesn’t necessarily capture all surface complexities. We acknowledge that future work will require power attenuation adjustment in the network or learning of the diffusion pattern based on material/texture to account for more diffused materials or a composite of materials. 
 
\section{Radio Frequency Background} \label{sec:app-radar}

\subsection{Radar Basics}
A mmWave radar works by transmitting a wireless
signal (chirp) and receiving back the reflections that come from various reflectors in the scene. In this case, the transmitter and receiver are collocated, meaning they are side by side.
It operates in the millimeter-wavelengths frequency bands at 77 GHz, and uses Frequency Modulated Continuous Wave (FMCW) and antenna arrays to help resolve spatial ambiguity. 
To resolve range ambiguity, the received chirp is multiplied with the conjugate of the transmitted chirp and can be expressed as a complex function:
\begin{equation}
\label{eq:fmcw-suppl}
s(t) =  A \cdot e^{-j 2 \pi(f+kt)   d /c}  = A \cdot e^{-(j 2\pi k \tau )t} \cdot e^{-j 2\pi f \tau}
\end{equation}
where $A$ is the signal amplitude, $d$ is the round-trip propagation distance, $c$ is the speed of light, $\tau=d/c$ is the round-trip delay, $f$ is the starting frequency, and $k$ is the chirp slope. For multiple reflectors, we simply receive the linear combination of all the reflections.

\subsection{Free-space Power Decay}\label{sec:app-radar-pwr}
In free space, the power of a radio frequency signal is inversely proportional to the square of the distance it travels due to spherical spreading. 
We consider the round-trip path, where the signal travels from the transmitter to a point in space and then reflects back to the receiver, then decay is even more pronounced~\cite{richards2010principles}. 
This is known as \textit{round-trip free-space path loss}, and the \textit{power} decay factor reflected from distance \( u \) is given by:
\begin{equation}
\label{eq:reflection_dist}
P_{\text{rx}} \propto \frac{1}{(4\pi u)^4} P_{\text{tx}}
\end{equation}
This expression accounts for two instances of inverse-square spreading: one during transmission to the point and another during reflection back to the receiver. 
As such, the received power decreases proportionally to \( 1/u^4 \) (amplitude would be by a factor of \( 1/u^2 \)).

\subsection{Distance}\label{sec:app-radar-nf-ff}
Radio frequency signals are modeled differently based on the distance the transmitter and receiver are relative to the reflectors in the scene. 
Commonly, this is referred to as near-field (when the objects are in much closer to the transmitter/receiver pair compared to the wavelength of the radio frequency signal) and far-field (when the objects are much further compared to the wavelength). In our system, the object is much closer to the transmitter and receiver, as compared to the wavelength, meaning when the signals from the antenna array meet the object, the waves cannot be modeled as parallel. In other words, the direction of the RF signal from different parts of the antenna array will have vastly different directions. As compared to when the object is very far, then the waves can be modeled as parallel to each other~\cite{richards2010principles}, which significantly simplifies the radar heatmap reconstruction. This is because when the waves are assumed to be parallel we can reuse filter weights, and use Fourier Transforms to speed up the computation time~\cite{andersson2012fast}. 
On the other hand, in the near-field, we must use more accurate reconstruction methods (e.g. Matched Filter) which significantly increases the computational complexity of the system.

\subsection{Differentiable Matched Filter \& Signal Tracing}
\label{sec:app-mf}
We have the forward path of the Matched Filter:
\begin{equation*} 
P(\mathbf{x}_j) = \left\lVert \sum_{i = 1}^{N_{\text{ant}}} \sum_{t} s(i, t) \cdot e^{j 2\pi k \tau_i t} \cdot e^{j 2\pi f \tau_i} \right\rVert, \mathbf{x}_j \in \Omega_\text{pts},
\end{equation*}
The backpropagated gradient to the signal $s(i, t)$ is given by:
\begin{equation}
    \frac{\partial L}{\partial s(i,t)} = \sum_{\mathbf{x}_j \in \Omega_\text{pts}} \frac{1}{P(\mathbf{x})} \cdot \frac{\partial L}{\partial P(\mathbf{x})} \cdot e^{-j 2\pi k \tau_i t} \cdot e^{-j 2\pi f \tau_i}
\end{equation}

\noindent Signal Tracing forward path is defined by,
\begin{equation*} 
s(i, t) = \sum_{\mathbf{x}_j \in \Omega_\text{pts}} A_{\text{rx}}(\mathbf{x}_j)\, e^{-j 2\pi k \tau_{j} t} \, e^{-j 2\pi f \tau_{j}},
\end{equation*}
The backpropagated gradient to the amplitude $A_{\text{rx}}(\mathbf{x}_j)$ is given by:
\begin{equation} 
\frac{\partial L}{\partial A_{\text{rx}}(\mathbf{x}_j)}=  \sum_{i = 1}^{N_{\text{ant}}} \sum_{t} \frac{\partial L}{\partial s(i, t)} \cdot e^{j 2\pi k \tau_i t} \cdot e^{j 2\pi f \tau_i}
\end{equation}

\section{Future Work \& Discussion}
\label{sec:app-discussion} 
\noindent\textbf{Materials} While we present some initial results of non-metal reconstruction, our system currently shows results for objects that are primarily metal. This means the reflections we expect from the inner surface we are reconstructing reflect very clearly through the box. On the other hand, for objects which are made of a material which reflects much weaker than the box itself, may produce too noisy of matched filter heatmaps for our system to properly reconstruct the inside, because of the more complex RF interactions. Dealing with this requires adjusting the radar reflection model as well as the transmittance model. Additionally, future work will look into objects made of a composition of different materials. 

\noindent \textbf{Radar Scanning Methods }If there was a point that existed on the surface of the object, which never reflect \textit{at all} back to the radar in any of the scans taken, then the network has no way of truly recreating a point in that location, because it has never received any information from that location. Future work is required to deal with unseen surfaces, and more comprehensive scanning methods. 

\noindent \textbf{Computational Complexity}  
The computational cost of RF rendering still remains significantly higher compared to vision-based rendering.   
For comparison to state-of-the-art radar processing, the matched-filter takes roughly 40 minutes to process each of the 36 ground-truth images at 1 mm resolution. 
Compared to existing neural RF imaging baseline GeRaF~\cite{lu2025geraf}, which report training times near 32 hours per scene, our pipeline offers a similar efficiency. 
As neural RF representations are an emerging field, we believe that optimizing 3D space-sampling and training efficiency is a vital direction for future work.

\end{document}